\pdfoutput=1

\documentclass[11pt]{article}

\usepackage{acl}

\usepackage{pifont}  
\usepackage{fontawesome} 
\usepackage{array}
\usepackage{graphicx}
\usepackage{tcolorbox}
\usepackage{amsmath}
\usepackage{multirow}
\usepackage[normalem]{ulem}
\useunder{\uline}{\ul}{}

\usepackage{wasysym} 
\usepackage{times}
\usepackage{latexsym}
\usepackage{color}
\usepackage{tabularray}
\usepackage{amssymb} 
\usepackage{graphicx}
\usepackage[normalem]{ulem}
\usepackage[T1]{fontenc}

\usepackage[utf8]{inputenc}
\usepackage{tikz}

\usepackage[hang,flushmargin]{footmisc} 

\usepackage{listings}
\usepackage{fvextra}
\DefineVerbatimEnvironment{PromptBlock}{Verbatim}{
  breaklines=true,
  breakanywhere=true,
  fontsize=\small,
  baselinestretch=1,
  frame=single,
  xleftmargin=1em,
}
\usepackage{enumitem}

\newcommand{\fullcircle}{\CIRCLE}     
\newcommand{\halfcircle}{\LEFTCIRCLE} 
\newcommand{\emptycircle}{\Circle}    
\usepackage{microtype}

\title{TurnBench-MS: A Benchmark for Evaluating Multi-Turn, Multi-Step Reasoning in Large Language Models}

\author{
  Yiran Zhang\textsuperscript{1}, 
  Mo Wang\textsuperscript{1}, 
  Xiaoyang Li\textsuperscript{1},
   Kaixuan Ren\textsuperscript{1},\\
    \textbf{Chencheng Zhu}\textsuperscript{2},
  \textbf{Usman Naseem}\textsuperscript{1} \\
  \textsuperscript{1}Macquarie University 
  \textsuperscript{2}University of New South Wales \\
  \texttt{\{yiran.zhang,kaixuan.ren,usman.naseem\}@mq.edu.au}\\
}

\begin{document}
\maketitle

\begin{abstract}
Despite impressive advances in large language models (LLMs), existing benchmarks often focus on single-turn or single-step tasks, failing to capture the kind of iterative reasoning required in real-world settings. To address this limitation, we introduce \textbf{TurnBench}, a novel benchmark that evaluates multi-turn, multi-step reasoning through an interactive code-breaking task inspired by the "Turing Machine Board Game." In each episode, a model must uncover hidden logical or arithmetic rules by making sequential guesses, receiving structured feedback, and integrating clues across multiple rounds. This dynamic setup requires models to reason over time, adapt based on past information, and maintain consistency across steps—capabilities underexplored in current benchmarks. TurnBench includes two modes: \textit{Classic}, which tests standard reasoning, and \textit{Nightmare}, which introduces increased complexity and requires robust inferential chains. To support fine-grained analysis, we provide ground-truth annotations for intermediate reasoning steps. Our evaluation of state-of-the-art LLMs reveals significant gaps: the best model achieves 84\% accuracy in Classic mode, but performance drops to 18\% in Nightmare mode. In contrast, human participants achieve 100\% in both, underscoring the challenge TurnBench poses to current models. By incorporating feedback loops and hiding task rules, TurnBench reduces contamination risks and provides a rigorous testbed for diagnosing and advancing multi-step, multi-turn reasoning in LLMs\footnote{Our code and data is available at: \url{https://github.com/grantzyr/TurnBench-MS}}.
\end{abstract}

\section{Introduction}
\label{sec:introduction}

\begin{figure}[!t]
  \centering
  \includegraphics[width=\linewidth]{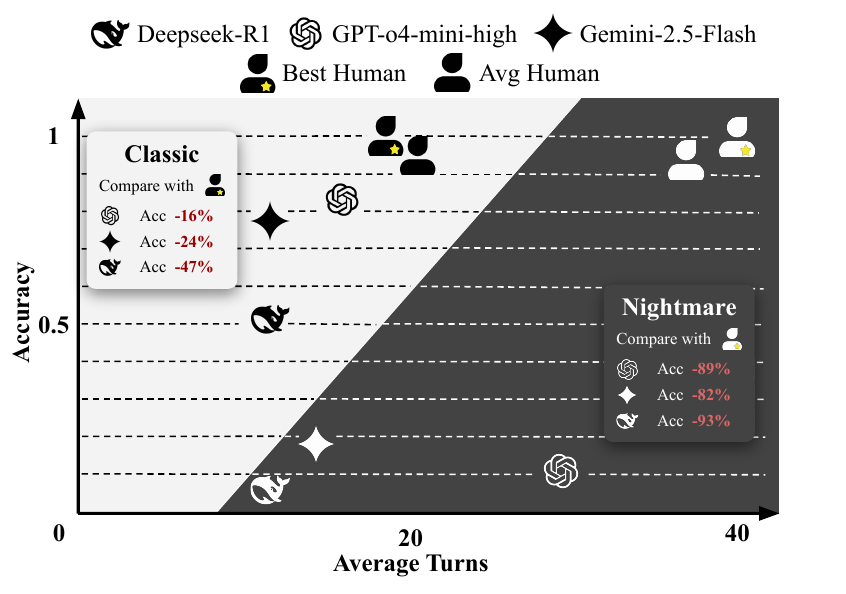}
  \caption{Accuracy versus average turns for leading LLMs and human evaluators (Best, Average) on TurnBench in both "Classic" and "Nightmare" modes. Insets show the relative accuracy drop of LLMs compared to the Best Human. Results highlight that LLMs remain substantially less accurate than humans, especially under the "Nightmare" setting, underscoring current limitations in complex multi-turn reasoning.}
  \label{fig:overall_image}
\end{figure}

Reasoning is central to human cognition and a key benchmark for evaluating the capabilities of artificial intelligence (AI) systems \cite{Wason1972-WASPOR-3, 10.1093/oxfordhb/9780199734689.013.0035}. In the context of large language models (LLMs), assessing reasoning ability is especially critical as these models are increasingly deployed in complex, real-world tasks. While a growing body of work has proposed datasets and evaluation methods for probing LLM reasoning \cite{zeng2024mr, wang2023can, welleck2022naturalprover}, such as Table \ref{tab:comparison_of_benchmarks}, significant gaps remain in how we measure and interpret this ability—particularly in multi-step, multi-turn settings.

\begin{figure*}[!t]
  \centering
  \includegraphics[width=0.99\textwidth]{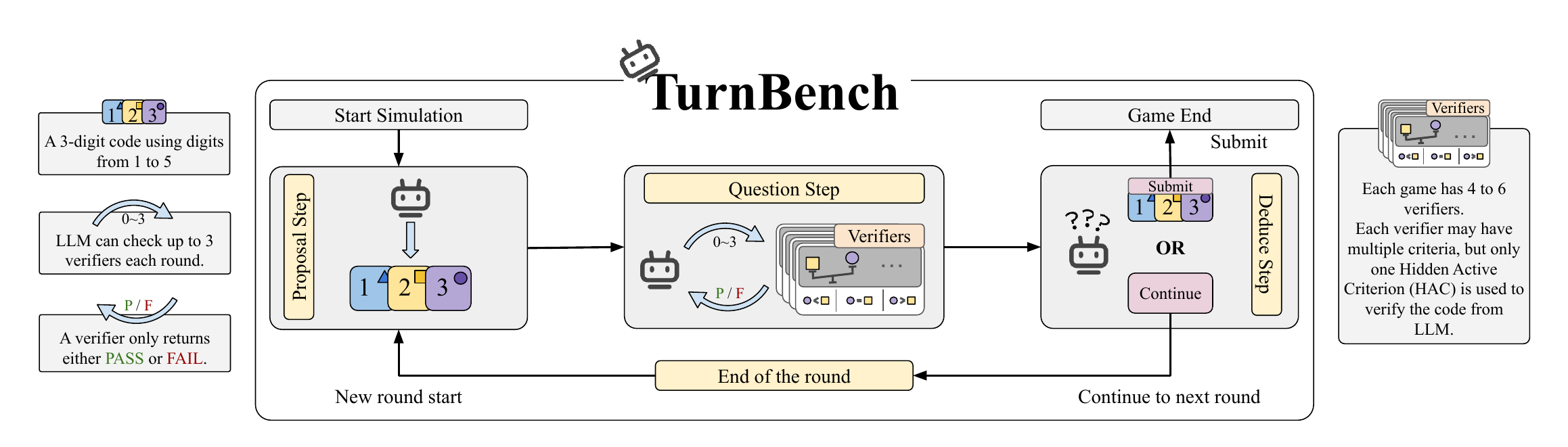}
  \caption{Overview of the TurnBench game framework.The LLM's objective is to deduce a secret 3-digit code composed of digits from 1 to 5. The game proceeds in iterative rounds, each comprising: 1) \textbf{Proposal Step}: The LLM submits a candidate 3-digit code. 2) \textbf{Question Step}: The LLM queries up to three verifiers, each providing Pass/Fail feedback based on its unique Hidden Active Criterion (HAC). 3) \textbf{Deduce Step}: The LLM analyzes the collective feedback to either \textbf{Submit} the final code if confident in its correctness, or 4) \textbf{Continue (End of the round} to the next round with a revised proposal. This iterative process continues until the LLM successfully deduces and submits the correct code.}
  \label{fig:flowchart}
\end{figure*}

First, most existing evaluations focus on single-turn or single-step reasoning tasks, overlooking the iterative and interactive nature of real-world problem-solving. Human reasoning often involves cycles of information gathering, hypothesis testing, and adaptation to feedback. This is especially true in scenarios where information is incomplete or distributed across multiple interactions. While recent benchmarks attempt to assess multi-step reasoning \cite{tang2025dsgbench, zeng2024mr}, they rarely simulate settings that require reasoning across multiple turns.

Second, current evaluation metrics typically emphasize final-answer correctness, with little insight into the model's intermediate reasoning process \cite{zhuang2023toolqa, hao2024llm}. As complex reasoning often admits multiple valid paths, simply scoring final outputs fails to distinguish between genuine inference and lucky guesses. Though some methods attempt process-level evaluation via manual annotation or automated proxies \cite{zeng2024mr, tang2025dsgbench}, these are limited by subjectivity and the absence of reliable ground truth for intermediate reasoning.

Third, data contamination poses a serious concern. Static benchmarks—often sourced from public datasets or templated questions—can overlap with pretraining corpora, making it difficult to disentangle memorization from actual reasoning \cite{yang2025dynamic, jain2024livecodebench, li2023starcoder}. This undermines the reliability of benchmark results and inflates perceived model performance.

To address these gaps, we introduce \textbf{TurnBench}, a novel benchmark designed to evaluate multi-turn, multi-step reasoning through an interactive code-breaking task inspired by the \textit{Turing Machine} board game. In this game, a model must uncover a hidden three-digit code by engaging in multiple rounds of interaction with logical verifiers. Each verifier is governed by a hidden rule; only one rule per verifier is active in a given instance. To succeed, the model must iteratively guess codes, select verifiers, analyze feedback, and gradually infer the underlying logical or arithmetic constraints—mirroring how humans perform exploratory reasoning.


\begin{table*}[htbp]
\centering
\resizebox{\linewidth}{!}{%
\begin{tabular}{lccccccc}
\hline
  Dataset & Multi-Turn & Multi-Step & No Knowledge & Ground true & Intermediate Eval & Reasoning & Domain \\
\hline
Avalonbench    & \fullcircle & \fullcircle & \fullcircle & \emptycircle & \emptycircle & \fullcircle & Game \\
Multi-LogiEval & \emptycircle    & \fullcircle & \emptycircle     & \fullcircle & \emptycircle & \fullcircle & Narrative \\
BoardgameQA    & \emptycircle     & \fullcircle & \halfcircle      & \fullcircle & \emptycircle & \halfcircle       & Game \\
MuSR           & \emptycircle    & \fullcircle & \emptycircle    & \fullcircle & \emptycircle & \fullcircle & Narrative \\
AIME 2024     & \emptycircle     & \fullcircle & \fullcircle & \fullcircle & \emptycircle & \fullcircle & Math \\
DSGBench       & \fullcircle & \emptycircle    & \emptycircle     & \emptycircle    & \halfcircle   & \halfcircle      & Game \\
MR-Ben         & \emptycircle     & \fullcircle & \fullcircle & \fullcircle & \fullcircle & \fullcircle & Science \\
LOGICGAME      & \emptycircle    & \fullcircle & \fullcircle & \fullcircle & \emptycircle    & \fullcircle & Game \\
MastermindEval & \fullcircle    & \fullcircle & \fullcircle & \fullcircle & \emptycircle    & \fullcircle & Game \\
LMAct     & \fullcircle & \fullcircle & \fullcircle & \emptycircle & \emptycircle & \fullcircle & Game \\
\hline
Ours           & \fullcircle & \fullcircle & \fullcircle & \fullcircle & \fullcircle & \fullcircle & Game \\

\hline
\end{tabular}
}
\caption{Comparison of multi-round reasoning benchmarks across six key criteria. A \fullcircle  
indicates presence of the feature, a \emptycircle  means no presence of the feature, and a \halfcircle indicates partial. The "Domain" column shows the task type of each benchmark.}
\label{tab:comparison_of_benchmarks}
\end{table*}

TurnBench explicitly addresses key shortcomings in existing benchmarks. First, it evaluates \textbf{multi-turn, multi-step reasoning} by requiring LLMs to adapt dynamically to feedback across multiple rounds and integrate partial clues to formulate and revise hypotheses over time. Second, it enables \textbf{process-level evaluation} through a rule-based mechanism that compares models’ intermediate inferences—i.e., their identification of active rules in each verifier—against ground truth, allowing structured analysis of reasoning steps beyond final answer correctness. Finally, TurnBench offers strong \textbf{contamination resistance} due to its dynamic rule configurations: even under fixed game setups, varying rule activations lead to distinct reasoning trajectories, minimizing the risk of data leakage from LLM pretraining corpora. Our work makes the following key contributions:
\begin{itemize}[noitemsep,leftmargin=*]
    \item We propose \textbf{TurnBench}, the novel benchmark designed to evaluate \textit{multi-turn, multi-step} reasoning in LLMs through dynamic, interactive tasks. TurnBench includes 540 game instances across two modes—\textit{Classic} and \textit{Nightmare}—with three difficulty levels each.
    
    \item We introduce a novel, automated evaluation method that leverages rule-based feedback to analyze intermediate reasoning steps, offering a grounded way to assess the internal thinking of LLMs.
    
    \item We benchmark a range of open-source and proprietary models, including GPT-o4-mini and Gemini-2.5-Flash, alongside human participants. Results show a significant performance gap between humans (100\%) and models (as low as 18\% in Nightmare mode), highlighting the challenge TurnBench presents (Figure \ref{fig:overall_image}).
    
    \item We release a new dataset comprising not only game settings and final answers, but also detailed interaction logs and reasoning steps for both models and humans, providing a valuable resource for future research.
\end{itemize}

\section{Related Work}

\paragraph{LLMs in Interactive Game Environments.} Recent work has explored the use of LLMs as agents in interactive games to assess their planning, reasoning, and decision-making capabilities across diverse domains such as board games, card games, and social deduction settings~\citep{schultz2024mastering, xu2023exploring, akata2023playing, light2023avalonbench, meta2022human, wang2023describe, zhuang2025pokerbench, tang2025dsgbench}. These benchmarks typically present the game state in textual or structured formats and prompt LLMs to make the next move using natural language generation. For instance, PokerBench~\citep{zhuang2025pokerbench} adopts classification-based decision scenarios, while AvalonBench~\citep{light2023avalonbench} and BALROG~\citep{paglieri2025balrogbenchmarkingagenticllm} evaluate agents through multi-turn, interactive gameplay. More recently, LMAct~\citep{ruoss2025lmactbenchmarkincontextimitation} and MastermindEval~\citep{golde2025mastermindevalsimplescalablereasoning} extend this line of work by framing game-based reasoning as multi-step, multi-turn interaction tasks. Unlike benchmarks focused on win rate or action legality, these settings emphasize the process of rule discovery and iterative hypothesis testing in constrained domains. Common evaluation metrics include win rate, legality of actions, strategy optimality, and task completion.

\paragraph{Benchmarks for Multi-Step and Logical Reasoning.} To more directly evaluate reasoning capabilities, recent studies have proposed benchmarks focused on multi-step logical and mathematical inference. LogiEval~\citep{patel2024multi}, Belief-R1~\citep{wilie2024belief}, MuSR~\citep{sprague2023musr}, and AIME~\citep{AIME2024} test multi-step reasoning in logical and mathematical domains, often revealing that even advanced LLMs struggle with deep inference. Complementary efforts such as CriticBench~\citep{lin2024criticbench} and MR-Ben~\citep{zeng2024mr} explore multi-round self-reflective prompting to enhance reasoning through critique and correction.


\paragraph{Rule-Based Inference and Tool-Augmented Reasoning.} Several benchmarks focus on rule-based or structured inference tasks. GridPuzzle and PuzzleEval~\citep{tyagi2024step} utilize logic grid puzzles, while ZebraLogic~\citep{lin2025zebralogic} frames reasoning as constraint satisfaction problems (CSPs). RuleArena~\citep{zhou2024rulearena} evaluates models on dynamic policy reasoning. Tool-augmented frameworks like LINC~\citep{olausson2023linc} and MATHSENSEI~\citep{das2024mathsensei} enable LLMs to perform formal reasoning through external tools. Meanwhile, self-reflection strategies such as Self-Refine~\citep{madaan2023self} and ReFlexion~\citep{shinn2023reflexion} allow models to iteratively revise incorrect or incomplete outputs via internal critique loops.

While the above efforts have made significant strides in evaluating LLM reasoning, several important gaps remain. First, few benchmarks explicitly evaluate \textit{multi-step reasoning across multiple interaction rounds}—a critical feature of real-world problem-solving. Most logic and tool-based tasks are static, single-shot evaluations that do not require models to gather and integrate information over time. Second, existing benchmarks often lack \textit{ground truth for intermediate reasoning steps}, limiting analysis to final-answer accuracy. This makes it difficult to determine whether a correct answer results from genuine reasoning or chance. Third, many datasets are vulnerable to \textit{data contamination} due to overlap with pretraining corpora. Finally, while game-based settings are promising, they rarely focus on rule-discovery and hypothesis refinement under feedback constraints.

TurnBench is designed to fill these gaps by offering a dynamic, interactive benchmark that simulates real-world multi-turn reasoning. It provides ground-truth for intermediate reasoning, minimizes contamination risk through dynamic rule configurations, and emphasizes logical consistency and rule inference across turns.

\section{TurnBench}

\subsection{Turing Machine Game Mechanics}
\label{lab:turing_machine_game_mechanics}
Turing Machine is a logic-based deduction game where the player's objective is to identify a unique three-digit code (digits 1–5), each digit associated with a distinct color (e.g., blue, yellow, purple). The challenge lies in interacting with a set of 4–6 verifiers, each governed by a single, hidden active criterion selected from a predefined rule pool. Players must deduce these hidden criteria and submit a code that satisfies all of them. The game rule and detailed example provided in Appendix \ref{apx:game_explanation}.

Each game unfolds in multiple rounds with four key phases: First, the player composes a proposed code (e.g., BLUE=2, YELLOW=4, PURPLE=3), which remains fixed for the current round. Next, the player queries up to three verifiers sequentially, each returning a binary judgment (PASS/FAIL) based on the verifier’s active rule. Then, using this feedback, the player can either attempt a final answer or skip and continue to the next round for further testing. The game ends once a final answer is submitted.

TurnBench supports two game modes: \textbf{Classic} and \textbf{Nightmare}, each with Easy, Medium, and Hard difficulty levels. In Classic mode, verifier responses correspond directly to the selected verifier’s criterion. In Nightmare mode, verifiers are secretly remapped; the player queries one verifier, but the response corresponds to another verifier’s logic, unknown to the player. This mapping must be deduced as part of the reasoning process. The mode details provided in Appendix \ref{apx:game_mode_explanation}.

\subsection{TurnBench Construction}

\subsubsection{Game Setups}

Each TurnBench game instance consists of a specific verifier combination, one hidden active rule per verifier, and the unique correct code. For Nightmare mode, each game additionally includes a fixed or dynamically generated hidden mapping between verifiers. We curated 270 Classic and 270 Nightmare game setups (90 per difficulty level), sourced from official materials\footnote{\url{https://www.turingmachine.info/}}. All setups are reproducible, and Nightmare mappings are pre-fixed or regenerated at runtime to reduce memorization risk.

\subsubsection{Verifier Design}
\begin{figure}[!t]
  \centering
  \includegraphics[width=\linewidth]{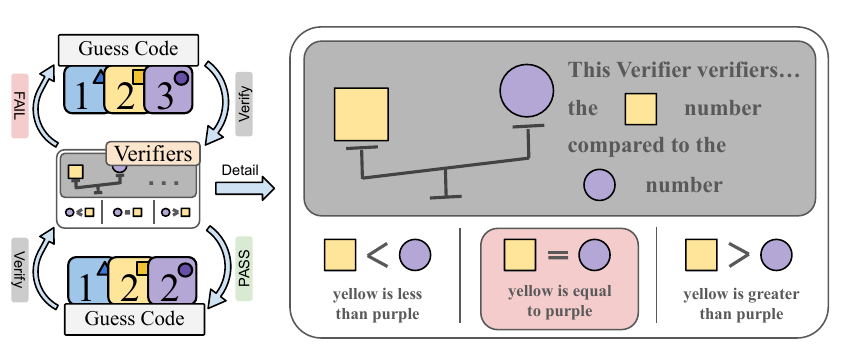}
  \caption{Example of verification process. This verifier (Right) compares the values assigned to yellow and purple. There are three possible criteria: less than, equal to, or greater than. The Hidden Active Criterion (HAC) (Red) represents the specific constraint activated by a verifier in a given game setup. When a tested code satisfies this criterion, the verifier returns "PASS"; otherwise, it returns "FAIL" (Left).}
  \label{fig:sample_verifier}
\end{figure}

Verifiers are central to TurnBench and encode simple numerical rules (e.g., Figure \ref{fig:sample_verifier}). We incorporate 48 official verifier types, each associated with 2 to 9 potential rules. Since the physical game's verifier logic isn't directly compatible with a simulation environment, we designed a Hidden Condition Selection Algorithm that selects one active rule per verifier to align with the game's design and balance.

\begin{figure*}[ht]
  \centering
  \includegraphics[width=\linewidth]{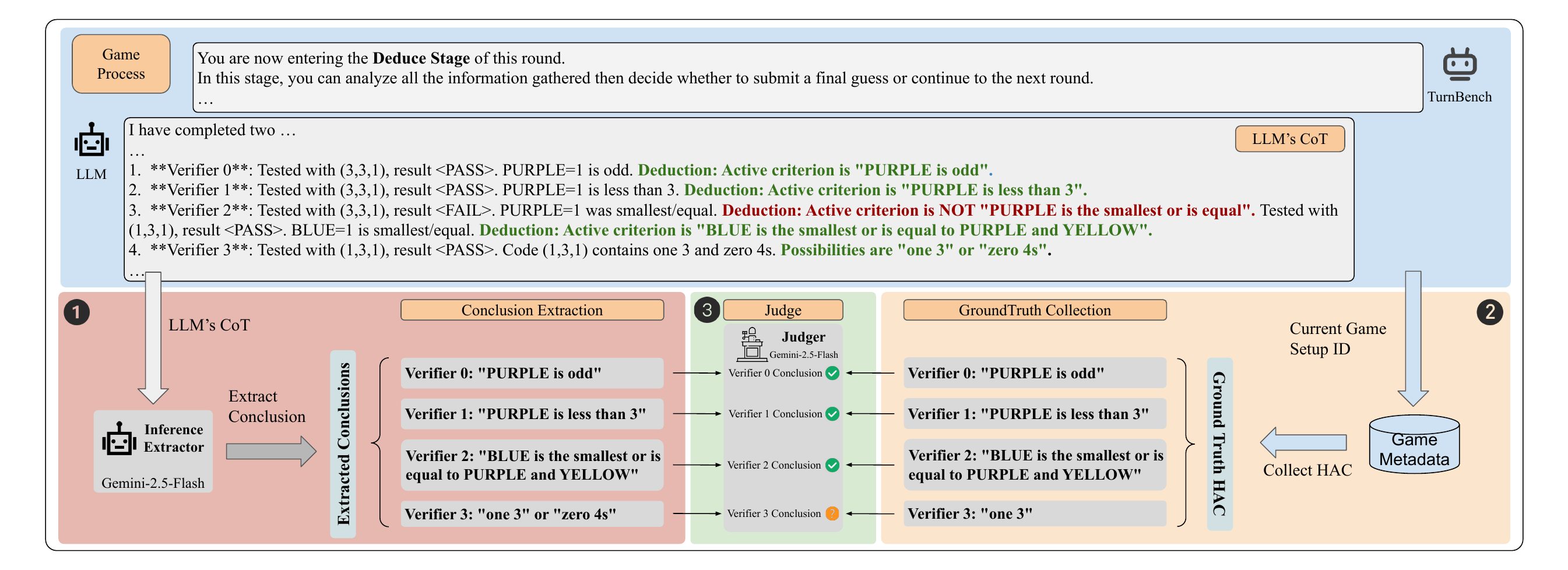}
  \caption{The Reasoning Process Evaluation Pipeline in TurnBench. This pipeline analyzes the LLM's Chain-of-Thought (CoT) generated as it deduces verifier properties during game process (blue). The evaluation proceeds in three steps: 1) \textbf{Inference Extraction} (red): The LLM's CoT, detailing its reasoning for each verifier's Hidden Active Criterion (HAC), is processed by the Inference Extractor. This yields "Extracted Conclusions" – the LLM's inferred HAC for each verifier. 2) \textbf{Ground Truth Collection} (orange): Simultaneously, the "Current Game Setup ID" is used to retrieve the definitive "Ground Truth HAC" for each verifier from the "Game Metadata". 3) \textbf{Judge} (green): The Judger then semantically compares the "Extracted Conclusions" from Step 1 with the corresponding "Ground Truth HAC" from Step 2. Each inferred HAC is categorized as: Correct (semantically equivalent to the ground truth), Incorrect (completely wrong), or Include (the conclusion contains the correct answer but is not yet fully refined to the precise ground truth).}
  \label{fig:thinking_process_evaluation}
\end{figure*}

\subsubsection{LLM Interaction Flow}

At game start, the system presents the LLM with the full game context: background, rules, objective, and all verifier definitions. The model then interacts turn-by-turn as described in Section \ref{lab:turing_machine_game_mechanics} (e.g. Figure \ref{fig:flowchart}), adhering to a strict output protocol. In each round:

\begin{itemize}
    \item In the \textbf{Proposal} step, the LLM outputs a code in the format \texttt{<CHOICE>: BLUE=X, YELLOW=Y, PURPLE=Z}.
    \item In the \textbf{Verifier Query} step, it selects verifiers with \texttt{<CHOICE>: [num]}. Each verifier returns PASS or FAIL.
    \item In the \textbf{Deduce} step, the LLM either submits the code again using the same format as Proposal step or skips the round via \texttt{<CHOICE>: SKIP}.
    \item In Chain-of-Thought (CoT) mode, the LLM also outputs reasoning before decisions using \texttt{<REASONING>}.
\end{itemize}

If the LLM produces malformed output or illegal actions (e.g., invalid verifier ID), a retry mechanism prompts re-generation, while tracking error frequency. Detailed prompts and retry protocols are in the Appendix \ref{apx:prompts}.

\subsubsection{Evaluating Model Reasoning Process}
\label{sec:evaluation_method_of_thinking_process}


While existing benchmarks focus solely on final answers, TurnBench introduces an automated method for evaluating intermediate reasoning. Specifically, in Classic mode, a model's reasoning involves two phases: (1) inferring each verifier’s hidden criterion, and (2) using these to deduce the final code. Since both ground truths (criteria and final code) are known, we can semantically compare model inferences with them.

Our evaluation pipeline (Figure ~\ref{fig:thinking_process_evaluation}) involves two LLM-based components. First, an \textbf{Inference Extractor} (Gemini-2.5-Flash~\cite{gemini25flash}) parses the model's \texttt{<REASONING>} output to identify its explicit claim about a verifier’s hidden rule. Second, a \textbf{Judger}, also Gemini-2.5-Flash, compares the extracted rule to the ground truth and classifies it as: \textbf{Correct} (semantically equivalent), \textbf{Incorrect} (completely wrong or missing the correct rule), or \textbf{Include} (partial overlap with the ground truth).

We validated this automated process through manual inspection. Stratified sampling selected 120 outputs (5\% of total), prioritizing failed games for robustness. Manual checks revealed the inference extractor missed 13.7\% of applicable conclusions, but achieved 99.7\% precision. The Judger reached 99.4\% classification accuracy. These results confirm that TurnBench provides a reliable mechanism for process-level evaluation of LLM reasoning.

\begin{table*}[ht]
\resizebox{\linewidth}{!}{%
\begin{tabular}{ccccccccccccccccccc}
\hline
\multirow{3}{*}{Models} &  & \multicolumn{11}{c}{Average Accuracy} &  & \multicolumn{2}{c}{\multirow{2}{*}{Win Avg Turn}} &  & \multicolumn{2}{c}{\multirow{2}{*}{Win Avg VER}} \\ \cline{3-13}
 &  & \multicolumn{2}{c}{Total} &  & \multicolumn{2}{c}{Easy} &  & \multicolumn{2}{c}{Medium} &  & \multicolumn{2}{c}{Hard} &  & \multicolumn{2}{c}{} &  & \multicolumn{2}{c}{} \\ \cline{3-4} \cline{6-7} \cline{9-10} \cline{12-13} \cline{15-16} \cline{18-19} 
 &  & OA & CoT &  & OA & CoT &  & OA & CoT &  & OA & CoT &  & OA & CoT &  & OA & CoT \\ \hline
gpt-o4-mini-high (Thinking) &  & \textbf{0.64} & \textbf{0.84} &  & 0.80 & \textbf{0.93} &  & \textbf{0.73} & \textbf{0.93} &  & \textbf{0.40} & \textbf{0.67} &  & 15 & 16 &  & 7 & 6 \\
gemini-2.5-flash (Thinking) &  & \textbf{0.64} & 0.76 &  & \textbf{0.87} & 0.87 &  & \textbf{0.73} & \textbf{0.93} &  & 0.33 & 0.47 &  & 14 & 13 &  & \textbf{6} & 6 \\
deepseek-r1 (Thinking) &  & 0.49 & 0.53 &  & 0.80 & 0.80 &  & 0.33 & 0.53 &  & 0.33 & 0.27 &  & \textbf{12} & 14 &  & \textbf{6} & 6 \\
gpt-4.1 &  & {\ul 0.09} & {\ul 0.69} &  & {\ul 0.20} & {\ul 0.80} &  & 0.07 & {\ul 0.73} &  & 0.00 & {\ul 0.53} &  & 46 & 15 &  & 23 & 7 \\
llama-4-maverick &  & 0.04 & 0.36 &  & 0.13 & 0.60 &  & 0.00 & 0.40 &  & 0.00 & 0.07 &  & {\ul 32} & 18 &  & {\ul 12} & 8 \\
qwen-2.5-7b-instruct &  & 0.04 & 0.04 &  & 0.00 & 0.13 &  & {\ul 0.13} & 0.00 &  & 0.00 & 0.00 &  & 42 & 7 &  & 21 & {\ul \textbf{3}} \\
mistral-8b &  & 0.00 & 0.02 &  & 0.00 & 0.00 &  & 0.00 & 0.00 &  & 0.00 & 0.07 &  & - & {\ul \textbf{6}} &  & - & {\ul \textbf{3}} \\
llama-3.1-8b-instruct &  & 0.00 & 0.00 &  & 0.00 & 0.00 &  & 0.00 & 0.00 &  & 0.00 & 0.00 &  & - & - &  & - & - \\ \hline
Random Guess &  & \multicolumn{2}{c}{0.0085} &  & \multicolumn{2}{c}{0.0079} &  & \multicolumn{2}{c}{0.0098} &  & \multicolumn{2}{c}{0.0077} &  & \multicolumn{2}{c}{-} &  & \multicolumn{2}{c}{-} \\ \hline
Best Human &  & \multicolumn{2}{c}{1} &  & \multicolumn{2}{c}{1} &  & \multicolumn{2}{c}{1} &  & \multicolumn{2}{c}{1} &  & \multicolumn{2}{c}{18} &  & \multicolumn{2}{c}{8} \\
Human Average &  & \multicolumn{2}{c}{0.96} &  & \multicolumn{2}{c}{0.98} &  & \multicolumn{2}{c}{0.95} &  & \multicolumn{2}{c}{0.95} &  & \multicolumn{2}{c}{20} &  & \multicolumn{2}{c}{11} \\ \hline
\end{tabular}
}
\caption{Performance of different models on the Classic Game setting (45 Games). Metrics include total, easy, medium, and hard average accuracy, as well as average number of turns and average number of verifiers used in successfully won games. Fewer average turns and verifier uses in winning games suggest greater reasoning efficiency. Human and random guess baselines are included for comparison. We evaluated two prompting strategies: Only Answer (OA) and Chain of Thought (CoT). The \textbf{bold text} represents the best results in LLM, the \underline{underline text} represents the best-performing result in the non-thinking model. Full test results for 270 setups are in Table \ref{tab:classic_game_results_270}.}
\label{tab:classic_game_results}
\end{table*}

\section{Experiment}
\subsection{Experiment Setup}

To comprehensively explore the limitations of current large language models (LLMs) in multi-turn and multi-step reasoning tasks, we selected both commercial and widely-used open-source models for evaluation, employing different prompting methods. The commercial models include gemini-2.5-flash-preview-04-17 (thinking) \cite{gemini25flash}, gpt-o4-mini-high-0416 (thinking) \cite{openaio4}, and gpt-4.1-2025-04-14. The open-source models tested are deepseek-r1 (thinking) \cite{deepseekai2025deepseekr1incentivizingreasoningcapability}, llama-4-maverick \cite{meta_llama4_blog2025}, mistral-8b \cite{ministral2024_ministraux}, llama-3.1-8b-instruct \cite{grattafiori2024llama3herdmodels}, and qwen-2.5-7b-instruct \cite{qwen2.5}. We also evaluated two prompting strategies: Only Answer (OA) and Chain of Thought (CoT) \cite{wei2022chain}. All models were evaluated through their publicly available APIs.

To thoroughly test the reasoning abilities of the state-of-the-art models, all "Thinking" models had their reasoning effort set to “high.” Additionally, to compare the reasoning gap between the most advanced LLMs and humans, we invited five human participants with no prior experience with the game to take part in the experiment.

We evaluated two game modes: Classic and Nightmare. Each mode's scenarios were divided equally into three difficulty levels: easy, standard, and hard. For Classic mode, we constructed 270 benchmark scenarios (90 per difficulty). For Nightmare mode, 45 scenarios were selected (15 per difficulty). Human participants played 45 Classic mode games (15 per difficulty), with the Nightmare mode evaluation set matching the models'.

All models and human participants were tested under identical conditions, with the same task prompts and problem setups (Appendix \ref{apx:human_player_interface}). To ensure parity in information access, we developed a user interface for humans that displayed exactly the same text as the models saw at each step. Humans were also asked to record their reasoning and thought processes throughout. Accuracy is calculated using scikit-learn~\citep{pedregosa2018scikitlearnmachinelearningpython}.

To specifically analyze the impact of model size on performance, we conducted a targeted follow-up experiment. We sampled 45 game setups from the Classic mode (15 from each difficulty level) and evaluated two additional large-scale open-source models: Llama-3.3-70B-instruct \cite{grattafiori2024llama3herdmodels} and Mistral-3.2-small \cite{mistral2025small32}. This allows for a more direct comparison across a range of model parameters.

\subsection{Results and Findings}

\begin{tcolorbox}
\textbf{Finding 1:} LLMs significantly lag behind humans in multi-turn, multi-step reasoning.
\end{tcolorbox}

\begin{table*}[ht]
\resizebox{\linewidth}{!}{
\begin{tabular}{ccclclclccccc}
\hline
\multirow{2}{*}{} &  & \multicolumn{7}{c}{Average Accuracy (CoT)} &  & \multirow{2}{*}{Win Avg Turn (CoT)} &  & \multirow{2}{*}{Win Avg VER (CoT)} \\ \cline{3-9}
 &  & Total &  & Easy &  & Medium &  & Hard &  &  &  &  \\ \hline
gpt-o4-mini-high (Thinking) &  & 0.11 &  & \textbf{0.13} &  & 0.20 &  & 0.00 &  & 21 &  & 8 \\
gemini-2.5-flash (Thinking) &  & \textbf{0.18} &  & \textbf{013} &  & \textbf{0.27} &  & \textbf{0.13} &  & 16 &  & 8 \\
deepseek-r1 (Thinking) &  & 0.07 &  & 0.07 &  & 0.07 &  & 0.07 &  & \textbf{12} &  & \textbf{6} \\ \hline
Random Guess &  & 0.0076 &  & 0.0074 &  & 0.0079 &  & 0.0075 &  & - &  & - \\ \hline
Best Human &  & 1 &  & 1 &  & 1 &  & 1 &  & 40 &  & 20 \\
Human Average &  & 0.94 &  & 0.96 &  & 0.93 &  & 0.93 &  & 38 &  & 17 \\ \hline
\end{tabular}
}
\caption{Performance of different thinking models on the Nightmare Game setting (45 Games). Same metrics as Table \ref{tab:classic_game_results}, but exclusively featuring the Chain of Thought (CoT) prompting strategy. Compared to Classic mode, accuracy drops significantly. Human players maintain robust performance, while models struggle to generalize under this challenging scenario.}
\label{tab:nightmare_game_results}
\end{table*}

We analyzed overall performance using average accuracy metrics segmented by difficulty (overall, easy, medium, hard), as well as the average number of turns and verifier uses in games won successfully. Fewer turns and verifier uses indicate stronger reasoning ability.

First, we discuss Classic mode results (Table~\ref{tab:classic_game_results}). Smaller standard models struggled significantly despite understanding game rules and response format. They had difficulty leveraging verifier feedback to make effective inferences. Because TurnBench requires no external knowledge and relies solely on numerical rules, this suggests that complex reasoning needs models of a certain size and capacity.

Chain of Thought (CoT) prompting consistently improved performance across accuracy metrics and helped "Thinking" models as well. For example, the best-performing gpt-o4-mini-high increased its overall accuracy from 64\% (OA) to 84\% (CoT). Larger standard models also showed notable gains, e.g., gpt-4.1 rose from 9\%(OA) to 69\% (CoT), and llama-4 from 4\% to 36\%.


CoT prompting also helped models succeed with fewer turns and verifier uses (e.g., gpt-4.1) dropped from 41 to 15 turns and from 23 to 7 verifiers). However, "Thinking" models showed little difference between OA and CoT for turns and verifier use, possibly because they internally perform stepwise reasoning. CoT may mainly help them articulate their reasoning process more clearly and use it as memory for subsequent steps.

Despite improvements, a significant gap remains between LLMs and humans. The "Best Human" achieved 100\% accuracy across all metrics, whereas gpt-o4-mini-high (CoT) reached only 84\%. But models outperformed human in average turns (20) and verifier uses (11). Analysis of reasoning logs showed that while models sometimes integrated more clues, humans tended to take more turns (especially on hard tasks) but maintained near-perfect accuracy.

To further test limits, we compared "Thinking" models with CoT against humans in the more challenging Nightmare mode (Table~\ref{tab:nightmare_game_results}). All LLMs’ accuracy dropped drastically compared to Classic mode. For example, gpt-o4-mini-high fell from 84\%to 11\% overall, and failed completely on Hard difficulty (0 accuracy). The best performing gemini-2.5-flash only reached 13\%. Humans maintained extremely high performance, with the "Best Human" at 100\% accuracy and the average human still achieving 94.2\%.

These results clearly demonstrate that although "Thinking" models and CoT prompting improve performance, LLMs still lag far behind humans in complex multi-turn, multi-step reasoning tasks, especially under high difficulty. This highlights the substantial gap remaining between current models and human reasoning capabilities.

\begin{tcolorbox}
\textbf{Finding 2:} Once LLMs make a mistake in multi-turn reasoning, they struggle to recover.
\end{tcolorbox}

\begin{table*}[h]
\centering
\resizebox{\textwidth}{!}{%
\begin{tblr}{
  width = \linewidth,
  colspec = {Q[c]Q[c]Q[c]Q[c]Q[c]Q[c]Q[c]Q[c]Q[c]},
  hline{1} = {1pt}, 
  hline{2} = {0.8pt}, 
  hline{8} = {1pt},
  row{1} = {font=\bfseries}
}
& llama-3.1-8b 
& gemini-2.5-flash 
& gpt-o4-mini-high 
& gpt-4.1 
& mistral-8b 
& llama-4-maverick 
& qwen-2.5-7b 
& deepseek-r1 \\

Initial verifier errors & 368 & 96 & \textbf{66} & 141 & 255 & 142 & 318 & 144 \\

Persistence of initial errors (\%) & 89.94 & 91.67 & \textbf{53.03} & 86.52 & 90.20 & 63.38 & 99.06 & 93.06 \\

Ended with no final conclusion (\%) & 74.18 & 71.87 & \textbf{34.85} & 54.61 & 53.33 & 45.77 & 96.23 & 86.11 \\

Next-turn still incorrect (\%) & 17.66 & 19.79 & 27.27 & 33.33 & 38.82 & 25.35 & \textbf{3.14} & 6.94 \\

Success despite persistent errors (\%) & 1.08 & 12.72 & \textbf{32.14} & 13.41 & 0.66 & 8.11 & 0.54 & 7.87 \\

Success when no / fixed errors (\%) & 1.75 & \textbf{95.34} & 87.55 & 84.57 & 3.13 & 41.75 & 8.00 & 90.56 \\
\end{tblr}
}
\caption{
Comparison of large language models on their ability to handle verifier errors during multi-turn reasoning. Metrics include the number of initial verifier errors, error persistence rate, likelihood of remaining incorrect in the next turn, and task success rates depending on error persistence or correction.}
\label{tab:verifier-errors}
\end{table*}

In this experiment, we conducted an in-depth analysis of the persistence and evolution of error states in LLMs during multi-turn reasoning. This analysis is based on the thinking process evaluation method described in Section~\ref{sec:evaluation_method_of_thinking_process}. The results clearly demonstrate that in complex multi-turn reasoning chains, once current LLMs make an initial error, they tend to "lose their way" and struggle to recover autonomously, significantly reducing their final task success rate.

\paragraph{Path Divergence after Initial Errors.}  
Using a Sankey diagram (Figure~\ref{fig:error_path}), we tracked model behavior following the First Incorrect Conclusion (FIC). The diagram shows that a large proportion of error paths led directly to "No Subsequent Conclusion," indicating that models often cease reasoning along that path after an initial mistake; indeed, as detailed in Table \ref{tab:verifier-errors}, the rate of "Ended with no final conclusion" ranged from 34.85\% (gpt-o4-mini-high) to a striking 96.23\% (qwen-2.5-7b) across various models. Another substantial fraction continued producing incorrect conclusions; for instance, the "Next-turn still incorrect" metric in the same table varied from 3.14\% (qwen-2.5-7b) to 38.82\% (mistral-8b). In contrast, paths that quickly shifted to "Include" or "Correct" were relatively rare. Examining how these paths evolved to the Final Conclusion State Before Submission (CSBS), we found that those with either "No Subsequent Conclusion" or "Subsequent Incorrect Conclusion" overwhelmingly ended in an incorrect final conclusion. Consequently, these error paths almost always resulted in "Game Lost." Only a small minority of paths that rapidly adjusted to correct or partially correct conclusions after the first error were associated with a higher likelihood of "Game Won." This divergence visually confirms that after the first mistake, models rarely self-correct and tend either to halt reasoning or perpetuate errors—an initial indication of "losing their way." The detailed definitions and examples are provided in Appendix \ref{apx:definitions_and_examples}.

\begin{figure}[h!]
  \centering
  \includegraphics[width=0.48\textwidth]{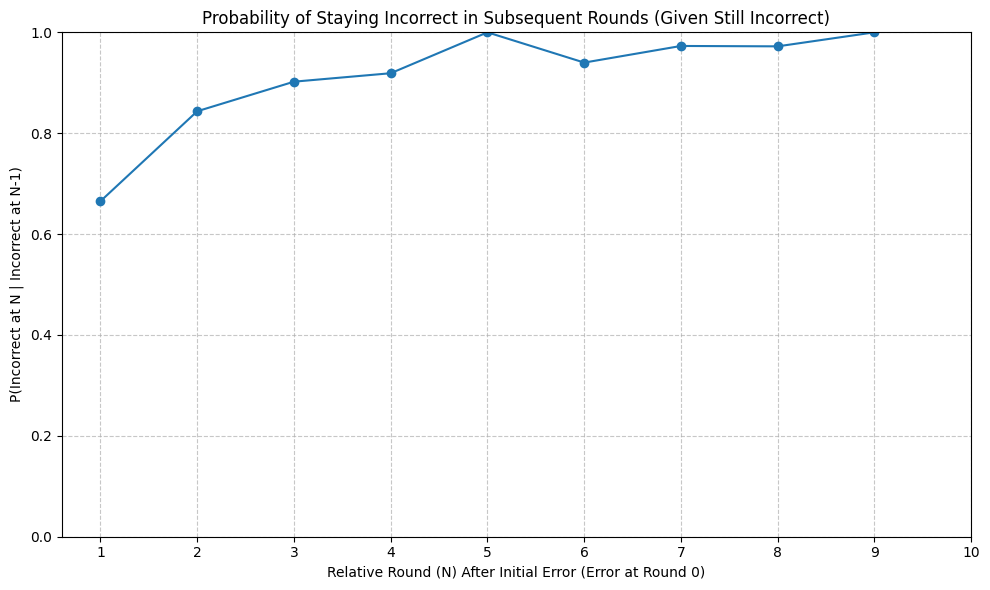}
  \caption{Probability of a model remaining incorrect in each subsequent round after its initial error, conditioned on it being incorrect in the previous round. The likelihood of continuing in an incorrect state increases with each turn, approaching near certainty beyond the fifth round. This trend highlights the models’ limited capacity for self-correction once they enter an error state.}
  \label{fig:staying_incorrect}
\end{figure}

\paragraph{Solidification and Persistence of Error States.}  
To further investigate error dynamics, we analyzed model behavior after making an error. Error states proved extremely "sticky." Figure~\ref{fig:staying_incorrect} depicts the probability that a model continues to produce incorrect conclusions in subsequent relative rounds, given that it is currently incorrect. In the first relative round after the initial error (X=1), if the model outputs a conclusion, there is already approximately a 65--70\% chance it is incorrect. Alarmingly, this probability rises sharply with additional rounds, nearing 100\% by the fifth relative round. This suggests that once a model enters several consecutive rounds of incorrect reasoning, it almost completely loses the ability to break the error cycle.

\subsection{Further Analysis of Reasoning Failures}

To better understand the sources of the performance gaps and the persistence of errors documented in Findings 1 and 2, we conducted additional analyses on both the qualitative nature of LLM errors and the quantitative effect of model scale.

\subsubsection{Qualitative Error Analysis}

We examined failure cases across different models and difficulty levels to investigate the causes of the performance gaps identified in Finding 1 and the error persistence highlighted in Finding 2. This analysis revealed five recurring categories of reasoning errors that illustrate the challenges LLMs face in multi-turn, multi-step tasks. Illustrative examples are provided in Appendix \ref{apx:example_of_llm_error_types}.

\paragraph{Reasoning Biases.} These errors occur when a model attempts to make inferences from insufficient evidence. In our setting, verifier rules often share implicit relationships. Some models, seeking efficiency, try to deduce the hidden criteria of untested verifiers based on already verified conditions or by exploiting the uniqueness of the final answer. While such "shortcut" strategies can be effective, they frequently lead to mistakes when models fail to account for alternative logical paths or conditions.

\paragraph{Task Misconception.} A small subset of models fail to fully grasp the core game mechanics or specific requirements, leading to misinterpretation of clues and ultimately to erroneous conclusions. This type of error is more common in smaller models, and is rarely observed in larger ones, pointing to limitations in complex instruction following and sustained task comprehension.

\paragraph{Information Hallucination.} Here the model fabricates unverified information or erroneously assumes that a verification step has been completed. For instance, a model may conclude that a certain verifier’s criterion is active simply because its current solution happens to satisfy it, despite never explicitly testing that verifier. This highlights LLMs’ difficulties in managing uncertainty and preserving factual correctness.

\paragraph{Overconfident Inference.} In some cases, models display unwarranted certainty in their final predictions even without completing all required verification steps. They submit premature answers based on partial or speculative reasoning, revealing a tendency toward overconfidence.

\paragraph{Long-term Memory Decay.} In extended multi-turn reasoning, certain models struggle to retain and integrate crucial information across turns. This often results in forgotten or confused clues from earlier steps, which compromises downstream reasoning. Such cases point to ongoing challenges in maintaining contextual coherence and reliable recall in multi-turn interactions.

\subsubsection{The Impact of Model Scale}

Our main experiments (Table \ref{tab:classic_game_results}) showed a stark contrast in performance, particularly with smaller open-source models failing completely. To systematically investigate this, we analyzed the performance of several open-source models across a spectrum of sizes on a representative subset of 45 Classic mode games.

\begin{table}[ht]
\resizebox{\linewidth}{!}{
\begin{tabular}{lclc}
\hline
Model & Model Size &  & Avg Acc \\ \hline
deepseek-r1 (Thinking) (MoE) & 671B (37B) &  & 0.53 \\
llama-4-maverick (MoE) & 400B (17B) &  & 0.36 \\
llama-3.3-70b-instruct & 70B &  & 0.36 \\
mistral 3.2 small & 24B &  & 0.27 \\
llama-3.1-8b-instruct & 8B &  & 0.00 \\
mistral-8b & 8B &  & 0.02 \\
qwen-2.5-7b-instruct & 7B &  & 0.04 \\ \hline
\end{tabular}
}
\caption{A Comparative Analysis of Large Language Models at Various Scales.}
\label{tab:model_size_performance}
\end{table}

The results (Table \ref{tab:model_size_performance}) reveal a strong positive correlation between model size and accuracy on TurnBench. Our key observations are as follows:

\paragraph{Significant Performance Gap for Smaller Models} Models in the 7B–8B parameter range consistently achieve near-zero accuracy (0–4\%). This suggests that smaller LLMs largely lack the capacity for the complex, multi-turn reasoning required by TurnBench, possibly due to limitations in maintaining complex contexts, performing intricate deductions, or recovering from early errors.

\paragraph{Improved Performance with Increasing Scale} As model size increases to 24B (Mistral 3.2 Small) and 70B (Llama 3.3 70B), we observe a substantial improvement in average accuracy, reaching 27\% and 36\%, respectively. This indicates that larger models are better equipped to handle the game's complexities, benefiting from increased capacity for reasoning, memory, and instruction following.

\paragraph{Effectiveness of Large-Scale MoE Architectures} The Mixture-of-Experts (MoE) models, deepseek-r1 and llama-4-maverick, achieve high accuracies of 53\% and 36\%. Notably, deepseek-r1, with its massive 671B total parameters, is the top performer among this set, demonstrating that large-scale architectures, particularly those leveraging MoE, possess superior capabilities for this type of reasoning task.

This analysis supports the conclusion that scaling effects play a crucial role in LLM performance on complex, multi-turn reasoning tasks. It provides deeper insight into the current limitations of smaller models and underscores the benefits of larger model architectures for tackling such challenges.

\section{Conclusion}
In this paper, our investigation using TurnBench has clarified the capabilities and limitations of Large Language Models (LLMs) in multi-turn, multi-step reasoning. TurnBench addresses several key limitations of current benchmarks and offers an effective method for automatically analyzing the reasoning processes of LLMs. Using this framework, we evaluated multiple standard chat models and thinking models, uncovering key findings that highlight the limitations of existing models. In summary, TurnBench fills a gap in the evaluation of LLMs' multi-turn, multi-step reasoning capabilities and provides a novel solution for assessing model reasoning processes. We hope that our work will inspire further research into multi-turn reasoning.

\section*{Limitation}
Effectively and accurately measuring a model's thinking process has always been a challenge. The automated evaluation of model thinking processes proposed in this paper requires an evaluation framework built on rules, which lacks generality. Furthermore, using Gemini 2.5 Flash for model inference extraction still has certain limitations. Although the extracted results have shown high accuracy after manual evaluation, further research and optimization are still needed. In addition, since our work involves benchmarking large language models, there are potential risks such as the models producing biased outputs, which should be carefully considered when interpreting the results.

\bibliography{acl_latex}

\onecolumn
\appendix
\onecolumn
\label{sec:appendix}

\section{Game Explanation}
\label{apx:game_explanation}

This appendix provides detailed explanation includes Game mode explanation, Game Example,

\subsection{Mode Explanation}
\label{apx:game_mode_explanation}

In \textbf{Classic mode}, there's a direct, one-to-one correspondence between the verifier a player selects and the actual verifier that processes the code. If a player tests Verifier 1 and receives a "PASS" result, they can confidently conclude that their code passed Verifier 1's specific validation.

In \textbf{Nightmare mode}, a "verifier mapping" is generated at the start of the game. This mapping randomly shuffles the verifier numbers and their actual underlying verifiers. This means that if a player tests "Verifier 1" in Nightmare mode and receives a "PASS" result, that result could have come from any of the active verifiers, not necessarily the one labeled "Verifier 1."

In Nightmare mode, players must first infer the correspondence between the verifier results and the actual verifiers before they can even begin to deduce each verifier's hidden active criterion. Only after resolving this mapping can they then proceed to infer the rules and ultimately the correct three-digit code, adding an extra layer of complexity and uncertainty to the reasoning process.

\subsection{Game Rules}
\label{apx:game_rules}

\paragraph{The objective of the game} The goal is to infer a unique three-digit code, composed of numbers from 1 to 5, using the fewest possible rounds and verifier uses. While speed is a factor, accuracy in deduction is paramount.

\paragraph{Game Setup} In both Classic and Nightmare modes, each game involves 4-6 verifiers. Each verifier secretly activates a single criterion that restricts the final code. As mentioned, for Nightmare mode, a random verifier mapping is applied at the start, obscuring which labeled verifier corresponds to which actual criterion.

\paragraph{3-digit-code Rules} The three-digit code consists of numbers from 1 to 5, which can be repeated (e.g., 123, 221). Each digit is associated with a specific color: Blue for the first digit, Yellow for the second, and Purple for the third. This code is the unique combination that satisfies the hidden active criterion of all verifiers, and importantly, the active criteria of different verifiers are never conflicting.

\paragraph{Verifier Rules}
\begin{itemize}
    \item Active Criteria: Each verifier has multiple potential criteria (as shown in Figure 2), but only one is secretly chosen as "active" for a given game. Players are unaware of which criterion is active for any specific verifier.
    \item Validation Focus: When a player's proposed code is tested against a verifier, only whether the code satisfies the active criterion is checked. For instance, if an active criterion is "Blue = 3" and the player submits "211" (Blue=2, Yellow=1, Purple=1), a "FAIL" result only indicates that the Blue digit did not satisfy "Blue = 3." It does not imply that the Yellow or Purple digits are incorrect.
    \item Non-Overlapping Information: The active criteria selected across different verifiers in a game provide distinct information. For example, Verifier 1 will not have "Blue < 3" as its active criterion if Verifier 2 also has "Blue < 3" as its active criterion.
\end{itemize}

\paragraph{Game Structure (Rounds)} Before the game begins, players can see the details of the verifiers set up for the current game. Each game round consists of four stages, as show in Figure 1 in the paper:
\begin{itemize}
    \item Proposal: The player designs a three-digit code for testing in the current round. This code cannot be changed within the round.
    \item Question: The player can choose to test up to three verifiers sequentially with the proposed code. After each test, the player sees the result (PASS/FAIL) and can then decide whether to test another verifier or which one to test next. This stage can also be skipped entirely.
    \item Deduce: In this stage, the player can either submit a final three-digit code (which can be different from the proposed code in Stage 1) or choose to proceed to the next round. Once a code is submitted, the game concludes immediately, regardless of success or failure.
    \item End of Round: If the player chooses not to submit a final answer in the Deduce stage, the current round ends, and a new round begins.
\end{itemize}

\subsection{Game Example}
\label{apx:game_example}

To provide a concrete understanding of gameplay and human reasoning, below is a step-by-step example from a Classic mode game. Please note that this example illustrates a possible thought process, not necessarily the most optimal one, and is designed for clarity of rules. The correct answer for this game is "241" and the Hidden Active Criteria (HACs) for each verifier are marked.

\subsubsection{Game Setup}

\textbf{Final Answer}: "241"

\textbf{Verifiers}:
\begin{itemize}
    \item Verifier <1>: Verifies the YELLOW number compared to 4.
    \begin{itemize}
        \item Possible criteria: YELLOW is less than 4.
        \item Possible criteria: YELLOW is equal to 4. (HAC)
        \item Possible criteria: YELLOW is greater than 4.
    \end{itemize}
    \item Verifier <2>: Verifies the number of 3s in the code.
    \begin{itemize}
        \item Possible criteria: There are zero 3s in the code. (HAC)
        \item Possible criteria: There is one 3 in the code.
        \item Possible criteria: There are two 3s in the code.
        \item Possible criteria: There are three 3s in the code.
    \end{itemize}
    \item Verifier <3>: Verifies the BLUE number compared to the YELLOW number.
    \begin{itemize}
        \item Possible criteria: BLUE is less than YELLOW. (HAC)
        \item Possible criteria: BLUE is equal to YELLOW.
        \item Possible criteria: BLUE is greater than YELLOW.
    \end{itemize}
    \item Verifier <4>: Verifies which color's number is smaller than either of the others.
    \begin{itemize}
        \item Possible criteria: BLUE is smaller than PURPLE and YELLOW.
        \item Possible criteria: YELLOW is smaller than BLUE and PURPLE.
        \item Possible criteria: PURPLE is smaller than YELLOW and BLUE. (HAC)
    \end{itemize}
\end{itemize}

\subsubsection{Example Details}
\begin{PromptBlock}
ROUND 1 - Stage 1: Proposal

Player's Thinking: I want to test Verifiers 1, 2, and 3. For Verifier 1, "Yellow = 4" is a good test. For Verifier 2, by using two '3's (Blue and Purple), I can test if there are two 3s in the code. For Verifier 3, "Blue = 3" and "Yellow = 4" will help check "Blue < Yellow."
Proposed Code: 343

ROUND 1 - Stage 2: Question

Verifier Choose: 1, Result: PASS
Player's Thinking: Verifier 1 passed. I now know Yellow = 4.

Verifier Choose: 2, Result: FAIL
Player's Thinking: Verifier 2 failed. This means there are not two 3s in the code. Since I already know Yellow = 4 (from Verifier 1), the number of 3s can only be zero, one, or three. Given my proposed code 343, and knowing Yellow is 4, it's either zero 3s (if Blue and Purple aren't 3s) or one 3 (if one of Blue or Purple is a 3).

Verifier Choose: 3, Result: PASS
Player's Thinking: Verifier 3 passed. I now know Blue < Yellow. Since I already know Yellow = 4, Blue can be 1, 2, or 3.

ROUND 1 - Stage 3: Deduce

Player's Thinking: Based on results from Verifiers 1, 2, and 3: Yellow = 4, Blue can be 1, 2, or 3. The code has zero or one '3'. I need more information, especially about Verifier 4 and to narrow down Blue/Purple.
Deduce Choice: Continue to the next round.

ROUND 1 - Stage 4: End of Round.

ROUND 2 - Stage 1: Proposal

Player's Thinking: I need to test Verifier 2 further and Verifier 4. Using "345" will test if there's one '3' (for Verifier 2). For Verifier 4, if it passes, Blue (3) is the smallest. If it fails, then Purple (5) must be the smallest (since Blue < Yellow, Yellow cannot be the smallest).
Proposed Code: 345

ROUND 2 - Stage 2: Question

Verifier Choose: 2, Result: FAIL
Player's Thinking: Verifier 2 failed. Since I used '3' in the Blue position, and it failed, I now know there are zero 3s in the code.

Verifier Choose: 4, Result: FAIL
Player's Thinking: Verifier 4 failed. This means Purple is the smallest number (among Blue, Yellow, Purple).

Verifier Choose: SKIP
Player's Thinking: I have enough information now. I know all the HACs.

ROUND 2 - Stage 3: Deduce

Player's Thinking: Combining all results:
Yellow = 4 (from Round 1, Verifier 1)
Blue < Yellow (from Round 1, Verifier 3) => Blue can be 1, 2, or 3.
Zero 3s in the code (from Round 2, Verifier 2) => Blue cannot be 3. So, Blue can be 1 or 2.
Purple is the smallest number (from Round 2, Verifier 4) => Since Purple must be smaller than Blue (1 or 2) and Yellow (4), Purple must be 1, Blue must be 2.
Final Deduction: Blue = 2, Yellow = 4, Purple = 1.
Deduce Choice: Submit 241

Game Over. Result: Player wins the game!
\end{PromptBlock}

\section{Additional experimental results}

This appendix provides additional experimental results for classic game mode. We evaluated the reasoning performance of the model on the full dataset (270 Games) (Table \ref{tab:classic_game_results_270}).

\begin{table}[ht]
\resizebox{\linewidth}{!}{%
\begin{tabular}{ccccccccccccccccccc}
\hline
\multirow{3}{*}{Models} &  & \multicolumn{11}{c}{Average Accuracy} &  & \multicolumn{2}{c}{\multirow{2}{*}{Win Avg Turn}} &  & \multicolumn{2}{c}{\multirow{2}{*}{Win Avg VER}} \\ \cline{3-13}
 &  & \multicolumn{2}{c}{Total} &  & \multicolumn{2}{c}{Easy} &  & \multicolumn{2}{c}{Medium} &  & \multicolumn{2}{c}{Hard} &  & \multicolumn{2}{c}{} &  & \multicolumn{2}{c}{} \\ \cline{3-4} \cline{6-7} \cline{9-10} \cline{12-13} \cline{15-16} \cline{18-19} 
 &  & OA & CoT &  & OA & CoT &  & OA & CoT &  & OA & CoT &  & OA & CoT &  & OA & CoT \\ \hline
gpt-o4-mini-high (Thinking) &  & 0.578 & \textbf{0.815} &  & 0.756 & \textbf{0.933} &  & 0.7 & \textbf{0.9} &  & 0.278 & \textbf{0.611} &  & 16 & 16 &  & 7 & 7 \\
gemini-2.5-flash (Thinking) &  & \textbf{0.652} & 0.785 &  & \textbf{0.844} & 0.9 &  & \textbf{0.756} & 0.867 &  & \textbf{0.356} & 0.589 &  & 13 & 13 &  & \textbf{6} & 6 \\
deepseek-r1 (Thinking) &  & 0.511 & 0.63 &  & 0.733 & 0.756 &  & 0.511 & 0.722 &  & 0.289 & 0.411 &  & \textbf{12} & 13 &  & \textbf{6} & 6 \\
gpt-4.1 &  & 0.052 & {\ul 0.63} &  & 0.078 & {\ul 0.80} &  & 0.033 & {\ul 0.689} &  & {\ul 0.044} & {\ul 0.4} &  & 41 & 15 &  & 21 & 7 \\
llama-4-maverick &  & {\ul 0.07} & 0.326 &  & {\ul 0.133} & 0.444 &  & {\ul 0.056} & 0.367 &  & 0.022 & 0.167 &  & {\ul 28} & 17 &  & 12 & 8 \\
llama-3.1-8b-instruct &  & 0.007 & 0.015 &  & 0.011 & 0.022 &  & 0.011 & 0 &  & 0 & 0.022 &  & 23 & 13 &  & {\ul 11} & 6 \\
mistral-8b &  & 0 & 0.015 &  & 0 & 0.011 &  & 0 & 0.022 &  & 0 & 0.011 &  & - & 8 &  & - & 4 \\
qwen-2.5-7b-instruct &  & 0.015 & 0.022 &  & 0.011 & 0.067 &  & 0.022 & 0 &  & 0.011 & 0 &  & 34 & {\ul \textbf{6}} &  & 17 & {\ul \textbf{3}} \\ \hline
Random Guess &  & \multicolumn{2}{c}{0.0085} &  & \multicolumn{2}{c}{0.0079} &  & \multicolumn{2}{c}{0.0098} &  & \multicolumn{2}{c}{0.0077} &  & \multicolumn{2}{c}{-} &  & \multicolumn{2}{c}{-} \\ \hline
\end{tabular}
}
\caption{Performance of different models on the Classic Game setting (270 Games). Metrics include total, easy, medium, and hard average accuracy, as well as average number of turns and average number of verifiers used in successfully won games. Fewer average turns and verifier uses in winning games suggest greater reasoning efficiency. Human and random guess baselines are included for comparison. We evaluated two prompting strategies: Only Answer (OA) and Chain of Thought (CoT). The \textbf{bold text} represents the best results in LLM, the \underline{underline text} represents the best-performing result in the non-thinking model.}
\label{tab:classic_game_results_270}
\end{table}

\section{Definitions and Examples for Reasoning Path Evaluation}
\label{apx:definitions_and_examples}

\begin{figure}[ht]
  \centering
  \includegraphics[width=\textwidth]{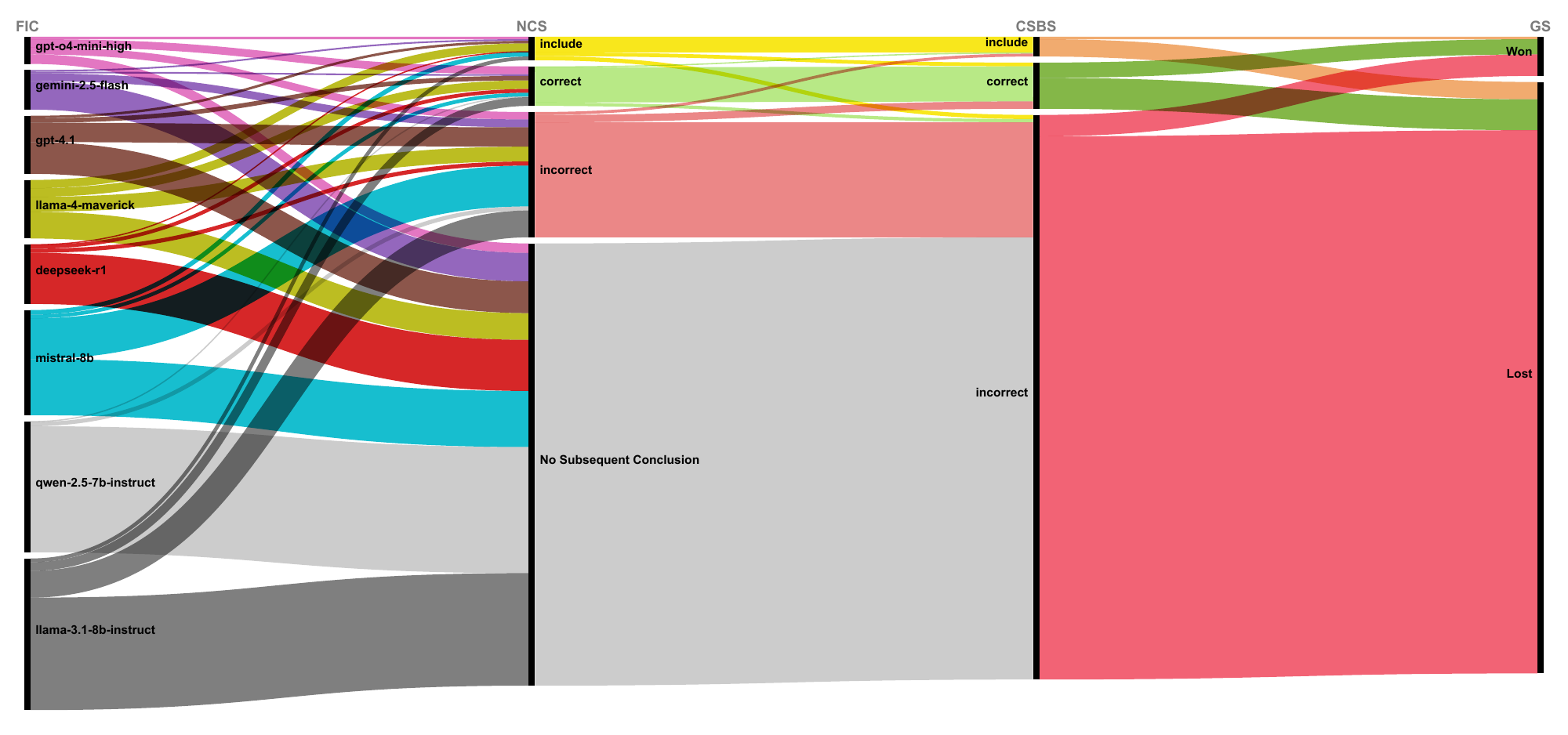}
  \caption{Flow analysis of verifier reasoning paths originating from a First Incorrect Conclusion (FIC). There are three subsequent stages: 1) \textbf{Next Conclusion Status (NCS)}: the outcome of the LLM’s next reasoning attempt on the same verifier. 2) \textbf{Conclusion Status Before Submit (CSBS)}: the final inferred status of the initially misjudged verifier before the LLM submitted its overall answer. 3) \textbf{Game Status (GS)}: the ultimate outcome (Won or Lost) of the game in which the FIC occurred. More detailed definitions and examples are provided in Appendix \ref{apx:definitions_and_examples}.}
  \label{fig:error_path}
\end{figure}

This appendix provides the detailed definitions and flow analysis figure (Figure \ref{fig:error_path}, and examples of the flow analysis of verifier reasoning paths originating from the First incorrect Conclusion.

\subsection{First Incorrect Conclusion (FIC)}
First Incorrect Conclusion(FIC) refers to the instances within a single game where the model makes an incorrect inference regarding a verifier's Hidden Active Criterion (HAC), HAC refer to Figure \ref{fig:sample_verifier}.

\subsection{Next Conclusion Status (NCS)}
The Next Conclusion Status (NCS) evaluates the outcome of an LLM's subsequent reasoning attempt on a specific verifier's Hypothesized Active Criteria (HAC) after a prior conclusion. Each attempt is classified into one of four categories, illustrated with examples referencing the verifier in Figure \ref{fig:sample_verifier}.

\textbf{Correct}: The inferred HAC is semantically equivalent to the ground truth rule. \textbf{Incorrect}: The inferred HAC is logically inconsistent with or fails to capture the ground truth. \textbf{Include}: The inferred HAC is a superset of or partially overlaps with the ground truth. It is consistent with the available evidence but is not yet the most specific or precise rule. \textbf{No Subsequent Conclusion}: The model does not revisit its reasoning for that specific verifier in any subsequent turn, effectively abandoning that line of inquiry.

To illustrate these categories, consider a verifier whose unknown ground truth HAC is \textbf{"Yellow = Purple"}. Suppose the model tests an input where Yellow=1 and Purple=3, causing the verifier to return FAIL (which indicates the HAC was not met).

If the model infers a broad but consistent rule like Yellow >= Purple, the status is \textbf{Include}. Because the conclusion includes the correct answer "Yellow = Purple" as a possibility. If the model infers Yellow > Purple, the status is \textbf{Incorrect}, as this hypothesis is missing the correct answer. A \textbf{Correct} status is achieved only when the model, perhaps by synthesizing evidence from multiple tests, precisely identifies the HAC as Yellow = Purple.

Finally, the \textbf{No Subsequent Conclusion} status is applied if the model, after making an initial incorrect inference about this verifier, fails to re-evaluate or correct its reasoning in later turns. This represents a failure of self-correction, where an erroneous path of reasoning is permanently dropped.

\subsection{Conclusion Status Before Submit (CSBS)}
Conclusion Status Before Submit (CSBS) is the final inferred status of the initially misjudged verifier before the LLM submitted its overall game answer. In this case, there are no \textbf{"No Subsequent Conclusion"} status anymore.

\subsection{Game Status (GS)}
Game Status (GS) is the ultimate outcome (Won or Lost) of the game in which the FIC occurred.

\section{Examples of LLM Error Types}
\label{apx:example_of_llm_error_types}

This appendix provides illustrative examples of the error categories discussed in Finding 3. The Hidden Active Criterion (HAC) refers to the secret rule a verifier uses to validate a code.

\subsection{Reasoning Biases Example}
\label{apx:finding3_examples_reasoning_biases}
The correct answer is 344 for this example.

Model's Known Information (from tested verifiers):
\begin{itemize}
    \item V1: BLUE is odd (B in {1, 3, 5}) (Comment: Correctly inferred HAC for V1)
    \item V0: YELLOW is ge 4 (Y in {4, 5}) (Comment: Correct path, true HAC is Y=4)
    \item V3: YELLOW ge PURPLE (Y ge P) (Comment: Correct path, true HAC is Y=P)
    \item V2: zero 1s (Digits must be from {2, 3, 4, 5}) (Comment: Correctly inferred HAC for V2)
    \item V4: Smallest is not PURPLE (P ge min(B, Y)) (Comment: Correct path, true HAC is Blue is smallest)
\end{itemize}

Model's Reasoning (Error Highlighted):
\begin{itemize}
    \item "Let's test the combinations: V3: Y=P, V4: B < Y\&P.
    \item Apply Y=P to the list: (3,4,4), (3,5,5), (5,4,4), (5,5,5).
    \item Apply V4: B < Y\&P to this new list:
    \begin{itemize}
        \item (3,4,4): 3 < 4 \& 3 < 4 (Yes) to (3,4,4) is a candidate.
        \item (3,5,5): 3 < 5 \& 3 < 5 (Yes) to (3,5,5) is a candidate.
        \item (5,4,4): 5 < 4 \& 5 < 4 (No).
        \item (5,5,5): 5 < 5 \& 5 < 5 (No).
    \end{itemize} 
    \item If V3 is Y=P and V4 is B < Y\&P, there are two possible codes: (3,4,4), (3,5,5). Not unique."
\end{itemize}

\textbf{Analysis}: The model fails to consider the constraint from V0 (YELLOW is ge 4) in combination with Y=P, leading it to miss the unique solution. If Y=4, then Y=P implies P=4, which makes (3,4,4) the unique answer. The model did not fully capture all potential conditions, thus incorrectly excluding the correct answer.

\subsection{Task Misconception}
\label{apx:finding3_examples_task_misconception}
The model submits code "123" to a verifier checking "how many 3s in the code." The verifier returns "PASS."

\textbf{Model's Incorrect Analysis}: "Since the code 123 passed Verifier <0>, we know that the code does not contain any 3s."

\textbf{Analysis}: This directly contradicts the expected logical interpretation (a 'PASS' implies the presence of a '3').

\subsection{Information Hallucination}
\label{apx:finding3_examples_information_hallucination}
In a game, the model only tested 3 out of 4 verifiers. Its deduction for the last verifier is:

\textbf{Model's Hallucination}: "The only remaining verifier is Verifier <3>, which checks the number of even numbers compared to the number of odd numbers. Since our code is BLUE=2, YELLOW=3, and PURPLE=4, we have 2 even numbers (2, 4) and 1 odd number (3). This satisfies the condition of having more even numbers than odd numbers. Therefore, the active criterion for Verifier <3> is that there are more even numbers."

\textbf{Analysis}: The model infers the active criterion for Verifier <3> based solely on its proposed code, without actually testing Verifier <3> or receiving feedback, hallucinating a verified state.

\subsection{Overconfident Inference}
\label{apx:finding3_examples_overconfident}
In a game where only 3 out of 4 verifiers were tested, the model concludes:

\textbf{Model's Overconfidence}: "Therefore, the code 324 satisfies all the known criteria and is likely to satisfy the hidden criteria of all verifiers."

\textbf{Analysis}: The model expresses high confidence ("likely to satisfy") despite not having complete information or having fully tested all verifiers, indicating a bias in self-assessment regarding the reliability of its deductions.

\subsection{Long-term Memory Decay}
\label{apx:finding3_examples_long_term_memory_decay}
Model tested Verifier 2 in the first round but forget in the second round

\textbf{Model's Deduction in Round 1}: "Verifier <2>: The result was <FAIL>, and since the sum of the digits (3+2+1 = 6) is a multiple of 3, the active criterion is either 'the sum is a multiple of 4' or 'the sum is a multiple of 5'"

\textbf{Model's Deduction in Round 2}: "We have not tested Verifier <2> yet. Our code's sum is 7, which is not a multiple of 3, 4, or 5, so we expect <FAIL> for Verifier <2> if tested." AND "the fact that we haven't tested Verifier <2>"

\textbf{Analysis}: The model clearly loses track of its prior interaction and deduction regarding Verifier <2>, indicating a failure to integrate past information into its current reasoning state.

\section{Prompts Used in Experiments}
\label{apx:prompts}

\subsection{Classic prompts with Only-Answer (OA) and Chain-of-Thought (CoT)}

\subsubsection{Classic system prompt}

\begin{PromptBlock}
You are participating in a competitive logic deduction game called Turing Machine.
Your goal is to win first place by deducing a secret 3-digit code with minimal rounds and verifier usage, but accuracy takes priority over speed.

Game Objective:
- Deduce the secret 3-digit code made up of digits 1-5.
- BLUE = first digit, YELLOW = second digit, PURPLE = third digit.
- Each digit can be 1, 2, 3, 4, or 5 (digits may repeat).
- The code is the ONLY combination that satisfies the active criterion of ALL chosen verifiers.

Game Structure (Rounds):
1. Proposal: Design a 3-digit code to test (format: BLUE=X, YELLOW=Y, PURPLE=Z, where X, Y, Z are digits from 1 to 5).
2. Question: Sequentially choose 0 to 3 verifiers to test your proposed code each round. After each selection, you will see the result, and then you can decide whether to select the next one.
3. Deduce: Based on verifier results, you can submit a final answer or continue to the next round.
4. End Round: If you didn't submit a final answer, a new round begins.

Verifier Rules:
- Each verifier checks ONE specific property (criterion) about the code.
- Each verifier has multiple potential criteria, but for each game, only ONE is secretly selected as 'active'. You don't know which criterion is active for any given verifier.
- Focus of Verification: When testing your code against a verifier, it exclusively evaluates it against its single, active criterion. The verifier completely ignores all other potential criteria, including its own inactive ones.
- PASS Condition: A verifier returns `<PASS>` if and only if your code satisfies this single active criterion.
- FAIL Condition: A verifier returns `<FAIL>` if and only if your code does not satisfy this single active criterion.
- Non-Overlapping Information: The active criteria selected across different verifiers for a game will provide distinct information.

Winning Strategy:
- It is possible to deduce the solution through joint reasoning, utilizing the combined results of multiple verifiers along with system rules such as the existence of a unique solution and the principle that no two verifiers offer redundant information.
- Only submit a final guess when you have either tested all verifiers and received <PASS> for each, or when your reasoning clearly proves your code satisfies all possible active verifier criteria. Accuracy takes priority over speed.

Current Game Setup:
{game_setup}
\end{PromptBlock}

\subsubsection{Classic proposal step prompt}

\paragraph{Classic - Proposal step - Step prompt -OA}
\begin{PromptBlock}
You are now entering the **Proposal Stage** of this round.

**Stage Purpose**:
In this stage, you need to compose a 3-digit code to help you to gather information from the verifiers. The code can NOT be changed in the subsequent stages of this round.

**3-digit code rules**:
- BLUE = first digit (X), YELLOW = second digit (Y), PURPLE = third digit (Z).  
- Each digit (X, Y, Z) can be 1, 2, 3, 4, or 5. Digits may repeat.

**Your Goal in This Stage**:
- Design a code that will test a specific hypothesis.
- Think about what a <PASS> or <FAIL> would tell you.
- Choose a code that lets you learn something meaningful from verifiers.

**What You Must Do Now**:
- Reply the code you want to use in this round with required response format. For example, <CHOICE>: BLUE=1, YELLOW=1, PURPLE=1
- DO NOT include any explanation, only follow the response format.

**Response format**:
<CHOICE>: BLUE=X, YELLOW=Y, PURPLE=Z
\end{PromptBlock}

\paragraph{Classic - Proposal step - Not valid format prompt - OA}
\begin{PromptBlock}
You did not follow the required response format. Please try again with same code.

**What You Must Do Now**:
- Reply the code you want to use in this round with required response format. For example, <PROPOSAL>: BLUE=1, YELLOW=1, PURPLE=1
- DO NOT include any explanation, only follow the response format.

**Response format**:
<CHOICE>: BLUE=[X], YELLOW=[Y], PURPLE=[Z]
\end{PromptBlock}

\paragraph{Classic - Proposal step - Step prompt - CoT}
\begin{PromptBlock}
You are now entering the **Proposal Stage** of this round.

**Stage Purpose**:
In this stage, you need to compose a 3-digit code to help you to gather information from the verifiers. The code can NOT be changed in the subsequent stages of this round.

**3-digit code rules**:
- BLUE = first digit (X), YELLOW = second digit (Y), PURPLE = third digit (Z).  
- Each digit (X, Y, Z) can be 1, 2, 3, 4, or 5. Digits may repeat.

**Your Goal in This Stage**:
- Design a code that will test a specific hypothesis.
- Think about what a <PASS> or <FAIL> would tell you.
- Choose a code that lets you learn something meaningful from verifiers.

**What You Must Do Now**:
- Reply the code you want to use in this round with required response format. For example, <PROPOSAL>: BLUE=1, YELLOW=1, PURPLE=1
- Explain your reasoning step by step with <REASONING> tag, then provide your code.

**Response format**:
<REASONING>: [Explain your reasoning step by step for choosing this code]
<CHOICE>: BLUE=[X], YELLOW=[Y], PURPLE=[Z]
\end{PromptBlock}

\paragraph{Classic - Proposal step - Not valid format prompt - CoT}
\begin{PromptBlock}
You did not follow the required response format. Please try again with same code.

**What You Must Do Now**:
- Reply the code you want to use in this round with required response format. For example, <PROPOSAL>: BLUE=1, YELLOW=1, PURPLE=1
- Explain your reasoning step by step with <REASONING> tag, then provide your code.

**Response format**:
<REASONING>: [Explain your reasoning step by step for choosing this code]
<CHOICE>: BLUE=[X], YELLOW=[Y], PURPLE=[Z]
\end{PromptBlock}

\subsubsection{Classic question step prompt}

\paragraph{Classic - Question step - First question prompt - OA}
\begin{PromptBlock}
You are now entering the **Verifier Questioning Stage** of this round.

**Current Verifiers**:
{verifier_descriptions}

**Stage Purpose**:
In this stage, you can test your proposed 3-digit code using verifiers. Each verifier checks one hidden criterion. Use the test results to gather information and refine your deduction.

**Verifier Rules Summary**:
- Each verifier has ONE secretly selected active criterion.
- <PASS> means your code satisfies this rule; <FAIL> means it does not.
- Active rules do NOT overlap between verifiers.

**Your Goal in This Stage**:
- Choosing verifiers is optional; testing 0 verifiers is allowed. If you want to choose the verifier, you must choose verifiers **one at a time**. After each result, you may decide whether to test another. You may choose to test 0 to 3 verifiers **in total** during this round.
- **Passing all tested verifiers does NOT mean the code is correct.** To win, your code must satisfy the hidden criterion of **all verifiers**, whether tested or not.

**What You Must Do Now**:
- If you want to choose a verifier to test your proposed code, reply with verifier_num after <CHOICE> tag, such as <CHOICE>: 1.
- If you want to skip verifier testing for this round, reply with SKIP after <CHOICE> tag, such as <CHOICE>: SKIP.
- DO NOT include any explanation, only follow the response format.

**Response format**:
<CHOICE>: [your_choice]
\end{PromptBlock}

\paragraph{Classic - Question step - Following questions prompt - OA}
\begin{PromptBlock}
You chose Verifier <{verifier_num}> and the result is <{verifier_result}>.

**What You Must Do Now**:
- If you want to choose the next verifier to test, reply with verifier_num after <CHOICE> tag, such as <CHOICE>: 1.
- If you want to skip verifier testing for this round, reply with SKIP after <CHOICE> tag, such as <CHOICE>: SKIP.
- DO NOT include any explanation, only follow the response format.

**Response format**:
<CHOICE>: [your_choice]
\end{PromptBlock}

\paragraph{Classic - Question step - Last question prompt - OA}
\begin{PromptBlock}
You chose Verifier <{verifier_num}> and the result is <{verifier_result}>.

You have now tested the maximum number of three verifiers for this round. The next stage is the Deduce Stage. If you want to test more verifiers or new code, you can choose SKIP during the Deduce Stage to move on to the next round.
\end{PromptBlock}

\paragraph{Classic - Question step - Not valid format prompt - OA}
\begin{PromptBlock}
You did not follow the required response format. Please try again with same choice.

**What You Must Do Now**:
- If you want to choose the next verifier to test, reply with verifier_num after <CHOICE> tag, such as <CHOICE>: 1.
- If you want to skip verifier testing for this round, reply with SKIP after <CHOICE> tag, such as <CHOICE>: SKIP.
- DO NOT include any explanation, only follow the response format.

**Response format**:
<CHOICE>: [your_choice]
\end{PromptBlock}

\paragraph{Classic - Question step - Not valid verifier choice prompt - OA}
\begin{PromptBlock}
You selected Verifier <{verifier_num}>, which is not a valid verifier number.

Please choose a valid verifier or SKIP to next stage.

**What You Must Do Now**:
- If you want to choose the next verifier to test, reply with verifier_num after <CHOICE> tag, such as <CHOICE>: 1.
- If you want to skip verifier testing for this round, reply with SKIP after <CHOICE> tag, such as <CHOICE>: SKIP.
- DO NOT include any explanation, only follow the response format.

**Response format**:
<CHOICE>: [your_choice]
\end{PromptBlock}

\paragraph{Classic - Question step - First question prompt - CoT}
\begin{PromptBlock}
You are now entering the **Verifier Questioning Stage** of this round.

Current Verifiers:
{verifier_descriptions}

**Stage Purpose**:
In this stage, you can test your proposed 3-digit code using verifiers. Each verifier checks one hidden criterion. Use the test results to gather information and refine your deduction.

**Verifier Rules Summary**:
- Each verifier has ONE secretly selected active criterion.
- <PASS> means your code satisfies this rule; <FAIL> means it does not.
- Active rules do NOT overlap between verifiers.

**Your Goal in This Stage**:
- Choosing verifiers is optional; testing 0 verifiers is allowed. If you want to choose the verifier, you must choose verifiers **one at a time**. After each result, you may decide whether to test another. You may choose to test 0 to 3 verifiers **in total** during this round.
- **Passing all tested verifiers does NOT mean the code is correct.** To win, your code must satisfy the hidden criterion of **all verifiers**, whether tested or not.

**What You Must Do Now**:
- If you want to choose a verifier to test your proposed code, reply with verifier_num after <CHOICE> tag, such as <CHOICE>: 1.
- If you want to skip verifier testing for this round, reply with SKIP after <CHOICE> tag, such as <CHOICE>: SKIP.
- Explain your reasoning step by step with <REASONING> tag, then provide your choice.

**Response format**:
<REASONING>: [Explain your reasoning step by step for choosing the verifier or skipping verifiers]
<CHOICE>: [your_choice]
\end{PromptBlock}

\paragraph{Classic - Question step - Following questions prompt - CoT}
\begin{PromptBlock}
You chose Verifier <{verifier_num}> and the result is <{verifier_result}>.

**What You Must Do Now**:
- If you want to choose the next verifier to test, reply with verifier_num after <CHOICE> tag, such as <CHOICE>: 1.
- If you want to skip verifier testing for this round, reply with SKIP after <CHOICE> tag, such as <CHOICE>: SKIP.
- Explain your reasoning step by step based on verifier result after <REASONING> tag, then provide your choice.

**Response format**:
<REASONING>: [Explain your reasoning step by step for choosing the verifier or skipping verifiers]
<CHOICE>: [your_choice]
\end{PromptBlock}

\paragraph{Classic - Question step - Last question prompt - CoT}
\begin{PromptBlock}
You chose Verifier <{verifier_num}> and the result is <{verifier_result}>.

You have now tested the maximum number of three verifiers for this round. The next stage is the Deduce Stage. If you want to test more verifiers or new code, you can choose SKIP during the Deduce Stage to move on to the next round.
\end{PromptBlock}

\paragraph{Classic - Question step - Not valid format prompt - CoT}
\begin{PromptBlock}
You did not follow the required response format. Please try again with same choice.

**What You Must Do Now**:
- If you want to choose the next verifier to test, reply with verifier_num after <CHOICE> tag, such as <CHOICE>: 1.
- If you want to skip verifier testing for this round, reply with SKIP after <CHOICE> tag, such as <CHOICE>: SKIP.
- Explain your reasoning step by step based on verifier result after <REASONING> tag, then provide your choice.

**Response format**:
<REASONING>: [Explain your reasoning step by step for choosing the verifier or skipping verifiers]
<CHOICE>: [your_choice]
\end{PromptBlock}

\paragraph{Classic - Question step - Not valid verifier choice prompt - CoT}
\begin{PromptBlock}
You selected Verifier <{verifier_num}>, which is not a valid verifier number.

Please choose a valid verifier or SKIP to next stage.

**What You Must Do Now**:
- If you want to choose the next verifier to test, reply with verifier_num after <CHOICE> tag, such as <CHOICE>: 1.
- If you want to skip verifier testing for this round, reply with SKIP after <CHOICE> tag, such as <CHOICE>: SKIP.
- Explain your reasoning step by step based on verifier result after <REASONING> tag, then provide your choice.

**Response format**:
<REASONING>: [Explain your reasoning step by step for choosing the verifier or skipping verifiers]
<CHOICE>: [your_choice]
\end{PromptBlock}

\subsubsection{Classic deduce step prompt}

\paragraph{Classic - Deduce step - Deduce result prompt}
\begin{PromptBlock}
The final guess is {submitted_code}. The answer is {answer}, the guess is {is_correct}.
\end{PromptBlock}

\paragraph{Classic - Deduce step - Step prompt - OA}
\begin{PromptBlock}
You are now entering the **Deduce Stage** of this round.

**Stage Purpose**:
In this stage, you can analyze all the information gathered then decide whether to continue to the next round or submit a final guess.

**Hint**:
- Passing all tested verifiers does not mean the code is correct if not all verifiers were tested. To be correct, the code must satisfy the hidden criteria of all verifiers, not just the ones you tested.
- You may choose not to test some verifiers if you can clearly reason that your code meets their requirements. But you must ensure every verifier is either tested and passed, or clearly justified through reasoning. Testing and passing only part of the verifiers is not enough if others are ignored.
- This stage **is not for testing**, you don't have to submit an answer; you can proceed to the next round to continue gathering information.
- Accuracy takes priority over speed. If you submit, the game will end, and an incorrect guess will result in immediate failure.

**Your Goal in This Stage**:
- Decide whether to submit the final guess or continue to the next round. Submit the final guess will end the game, continue to the next round will help you gather more information.
- Submission is not mandatory, you must make this decision based on your own reasoning.

**What You Must Do Now**:
- If you want to continue to the next round, reply with SKIP after <CHOICE> tag, such as <CHOICE>: SKIP
- If you want to submit a final guess to end the game, reply with BLUE=X, YELLOW=Y, PURPLE=Z after <CHOICE> tag, such as <CHOICE>: BLUE=1, YELLOW=1, PURPLE=1.
- DO NOT include any explanation, only follow the response format.

**Response format**:
<CHOICE>: [your_choice]
\end{PromptBlock}

\paragraph{Classic - Deduce step - Not valid format prompt - OA}
\begin{PromptBlock}
You did NOT follow the response format. Please try again.

**What You Must Do Now**:
- If you want to continue to the next round, reply with SKIP after <CHOICE> tag, such as <CHOICE>: SKIP
- If you want to submit a final guess to end the game, reply with BLUE=X, YELLOW=Y, PURPLE=Z after <CHOICE> tag, such as <CHOICE>: BLUE=1, YELLOW=1, PURPLE=1.
- DO NOT include any explanation, only follow the response format.

**Response format**:
<CHOICE>: [your_choice]
\end{PromptBlock}

\paragraph{Classic - Deduce step - Step prompt - CoT}
\begin{PromptBlock}
You are now entering the **Deduce Stage** of this round.

**Stage Purpose**:
In this stage, you can analyze all the information gathered then decide whether to submit a final guess or continue to the next round.

**Hint**:
- Passing all tested verifiers does not mean the code is correct if not all verifiers were tested. To be correct, the code must satisfy the hidden criteria of all verifiers, not just the ones you tested.
- You may choose not to test some verifiers if you can clearly reason that your code meets their requirements. But you must ensure every verifier is either tested and passed, or clearly justified through reasoning. Testing and passing only part of the verifiers is not enough if others are ignored.
- This stage **is not for testing**, you don't have to submit an answer; you can proceed to the next round to continue gathering information.
- Accuracy takes priority over speed. If you submit, the game will end, and an incorrect guess will result in immediate failure.

**Your Goal in This Stage**:
- Analysis all information gathered.
- Decide whether to submit the final guess or continue to the next round.
- Submission is not mandatory, you must make this decision based on your own reasoning.

**What You Must Do Now**:
- If you want to continue to the next round, reply with SKIP after <CHOICE> tag, such as <CHOICE>: SKIP
- If you want to submit a final guess to end the game, reply with BLUE=X, YELLOW=Y, PURPLE=Z after <CHOICE> tag, such as <CHOICE>: BLUE=1, YELLOW=1, PURPLE=1.
- Explain your reasoning step by step with <REASONING> tag, then provide your choice. If you want to submit a final guess, you must provide the reasons for not proceeding to the next round.

**Response format**:
<REASONING>: [Analysis and explain your reasoning step by step for continue to next round or submit final guess]
<CHOICE>: [your_choice]
\end{PromptBlock}

\paragraph{Classic - Deduce step - Not valid format prompt - CoT}
\begin{PromptBlock}
You did NOT follow the response format. Please try again.

**What You Must Do Now**:
- If you want to continue to the next round, reply with SKIP after <CHOICE> tag, such as <CHOICE>: SKIP
- If you want to submit a final guess to end the game, reply with BLUE=X, YELLOW=Y, PURPLE=Z after <CHOICE> tag, such as <CHOICE>: BLUE=1, YELLOW=1, PURPLE=1.
- Explain your reasoning step by step with <REASONING> tag, then provide your choice.

**Response format**:
<REASONING>: [Analysis and explain your reasoning step by step for submitting the final guess or continue to next round]
<CHOICE>: [your_choice]
\end{PromptBlock}

\subsection{Nightmare Prompts with Only-Answer (OA) and Chain-of-Thought (CoT)}

\subsubsection{Nightmare system prompt}

\begin{PromptBlock}
You are participating in a competitive logic deduction game called Turing Machine.
Your goal is to win first place by deducing a secret 3-digit code with minimal rounds and verifier usage, but accuracy takes priority over speed.

Game Objective:
- Deduce the secret 3-digit code made up of digits 1-5.
- BLUE = first digit, YELLOW = second digit, PURPLE = third digit.
- Each digit can be 1, 2, 3, 4, or 5 (digits may repeat).
- The code is the ONLY combination that satisfies the active criterion of ALL chosen verifiers.

Game Structure (Rounds):
1. Proposal: Design a 3-digit code to test (format: BLUE=X, YELLOW=Y, PURPLE=Z, where X, Y, Z are digits from 1 to 5).
2. Question: Sequentially choose 0 to 3 verifiers to test your proposed code each round. After each selection, you will see the result from an unknown verifier. The verifier identity will be hidden.
3. Deduce: Based on verifier results, you can submit a final answer or continue to the next round.
4. End Round: If you didn't submit a final answer, a new round begins.

Verifier Rules:
- Each verifier checks ONE specific property (criterion) about the code.
- Each verifier has multiple potential criteria, but for each game, only ONE is secretly selected as 'active'. You don't know which criterion is active for any given verifier.
- Focus of Verification: When testing your code against a verifier, it EXCLUSIVELY evaluates it against its SINGLE, ACTIVE criterion. The verifier completely ignores all other potential criteria, including its own inactive ones.
- In this game, you don’t know which Verifier’s result you’re actually seeing — the mapping between Verifiers and their displayed results is randomized and hidden from the player, though fixed for the entire game.
- PASS Condition: A verifier returns `<PASS>` if and only if your code satisfies the active criterion of the actual Verifier it is mapped to. For example, if Verifier 1 is secretly mapped to Verifier 2, then a <PASS> from Verifier 1 means your code met Verifier 2's hidden active rule.
- FAIL Condition: A verifier returns `<FAIL>` if and only if your code does not satisfy the active criterion of the actual Verifier it is mapped to. A <FAIL> simply means the mapped Verifier's rule was not met.
- Non-Overlapping Information: The active criteria selected across different verifiers for a game will provide distinct information. 

Winning Strategy:
- It is possible to deduce the solution through joint reasoning, utilizing the combined results of multiple verifiers along with system rules such as the existence of a unique solution and the principle that no two verifiers offer redundant information.
- One possible strategy is to carefully modify your code across multiple rounds and observe how each Verifier’s output changes. By analyzing the pattern of responses, you can infer the hidden mapping between Verifiers and their actual criteria.
- Only submit a final guess when you have either tested all verifiers and received <PASS> for each, or when your reasoning clearly proves your code satisfies all possible active verifier criteria. Accuracy takes priority over speed.

Current Game Setup:
{game_setup}
\end{PromptBlock}

\subsubsection{Nightmare proposal step prompt}

\paragraph{Nightmare - Proposal step - Step prompt - OA}
\begin{PromptBlock}
You are now entering the **Proposal Stage** of this round.

**Stage Purpose**:
In this stage, you need to compose a 3-digit code to help you to gather information from the verifiers. The code can NOT be changed in the subsequent stages of this round.

**3-digit code rules**:
- BLUE = first digit (X), YELLOW = second digit (Y), PURPLE = third digit (Z).  
- Each digit (X, Y, Z) can be 1, 2, 3, 4, or 5. Digits may repeat.

**Your Goal in This Stage**:
- Design a code that will test a specific hypothesis.
- Think about what a <PASS> or <FAIL> would tell you, but you don’t know which Verifier’s result you’re actually seeing — the mapping between Verifiers and their displayed results is randomized and hidden from the player, though fixed for the entire game.
- Choose a code that lets you learn something meaningful from verifiers.

**What You Must Do Now**:
- Reply the code you want to use in this round with required response format. For example, <CHOICE>: BLUE=1, YELLOW=1, PURPLE=1
- DO NOT include any explanation, only follow the response format.

**Response format**:
<CHOICE>: BLUE=X, YELLOW=Y, PURPLE=Z
\end{PromptBlock}

\paragraph{Nightmare - Proposal step - Not valid format prompt - OA}
\begin{PromptBlock}
You did not follow the required response format. Please try again with same code.

**What You Must Do Now**:
- Reply the code you want to use in this round with required response format. For example, <PROPOSAL>: BLUE=1, YELLOW=1, PURPLE=1
- DO NOT include any explanation, only follow the response format.

**Response format**:
<CHOICE>: BLUE=[X], YELLOW=[Y], PURPLE=[Z]
\end{PromptBlock}

\paragraph{Nightmare - Proposal step - Step prompt - CoT}
\begin{PromptBlock}
You are now entering the **Proposal Stage** of this round.

**Stage Purpose**:
In this stage, you need to compose a 3-digit code to help you to gather information from the verifiers. The code cannot be changed in the subsequent stages of this round.

**3-digit code rules**:
- BLUE = first digit (X), YELLOW = second digit (Y), PURPLE = third digit (Z).  
- Each digit (X, Y, Z) can be 1, 2, 3, 4, or 5. Digits may repeat.

**Your Goal in This Stage**:
- Design a code that will test a specific hypothesis.
- Think about what a <PASS> or <FAIL> would tell you, but you don’t know which Verifier’s result you’re actually seeing — the mapping between Verifiers and their displayed results is randomized and hidden from the player, though fixed for the entire game.
- Choose a code that lets you learn something meaningful from verifiers.

**What You Must Do Now**:
- Reply the code you want to use in this round with required response format. For example, <PROPOSAL>: BLUE=1, YELLOW=1, PURPLE=1
- Explain your reasoning step by step with <REASONING> tag, then provide your code.

**Response format**:
<REASONING>: [Explain your reasoning step by step for choosing this code]
<CHOICE>: BLUE=[X], YELLOW=[Y], PURPLE=[Z]
\end{PromptBlock}

\paragraph{Nightmare - Proposal step - Not valid format prompt - CoT}
\begin{PromptBlock}
You did not follow the required response format. Please try again with same code.

**What You Must Do Now**:
- Reply the code you want to use in this round with required response format. For example, <PROPOSAL>: BLUE=1, YELLOW=1, PURPLE=1
- Explain your reasoning step by step with <REASONING> tag, then provide your code.

**Response format**:
<REASONING>: [Explain your reasoning step by step for choosing this code]
<CHOICE>: BLUE=[X], YELLOW=[Y], PURPLE=[Z]
\end{PromptBlock}

\subsubsection{Nightmare question step prompt}

\paragraph{Nightmare - Question step - First question prompt - OA}
\begin{PromptBlock}
You are now entering the **Verifier Questioning Stage** of this round.

**Current Verifiers**:
{verifier_descriptions}

**Stage Purpose**:
In this stage, you can test your proposed 3-digit code using verifiers. Each verifier checks one hidden criterion. Use the test results to gather information and refine your deduction.

**Verifier Rules Summary**:
- Each verifier has ONE secretly selected active criterion.
- Each verifier shows results for a different, hidden verifier (the mapping is randomized but fixed for the entire game).
- <PASS> means your code satisfies the active criterion of the secretly mapped verifier. <FAIL> means your code does not satisfy that criterion.
- Active rules do NOT overlap between verifiers.

**Your Goal in This Stage**:
- Choosing verifiers is optional; testing 0 verifiers is allowed. If you want to choose the verifier, you must choose verifiers **one at a time**. After each result, you may decide whether to test another. You may choose to test 0 to 3 verifiers **in total** during this round.
- **Passing all tested verifiers does NOT mean the code is correct.** To win, your code must satisfy the hidden criterion of **all verifiers**, whether tested or not.

**What You Must Do Now**:
- If you want to choose a verifier to test your proposed code, reply with verifier_num after <CHOICE> tag, such as <CHOICE>: 1.
- If you want to skip verifier testing for this round, reply with SKIP after <CHOICE> tag, such as <CHOICE>: SKIP.
- DO NOT include any explanation, only follow the response format.

**Response format**:
<CHOICE>: [your_choice]
\end{PromptBlock}

\paragraph{Nightmare - Question step - Following questions prompt - OA}
\begin{PromptBlock}
You chose Verifier <{verifier_num}> and the result is <{verifier_result}>.

**Hint**:
- `<PASS>` means your code satisfies the active criterion of the actual Verifier it is mapped to. For example, if Verifier 1 is secretly mapped to Verifier 2, then a <PASS> from Verifier 1 means your code met Verifier 2's hidden active rule.
- `<FAIL>` means your code does not satisfy the active criterion of the actual Verifier it is mapped to. 

**What You Must Do Now**:
- If you want to choose the next verifier to test, reply with verifier_num after <CHOICE> tag, such as <CHOICE>: 1.
- If you want to skip verifier testing for this round, reply with SKIP after <CHOICE> tag, such as <CHOICE>: SKIP.
- DO NOT include any explanation, only follow the response format.

**Response format**:
<CHOICE>: [your_choice]
\end{PromptBlock}

\paragraph{Nightmare - Question step - Last question prompt - OA}
\begin{PromptBlock}
You chose Verifier <{verifier_num}> and the result is <{verifier_result}>.

You have now tested the maximum number of three verifiers for this round. The next stage is the Deduce Stage. If you want to test more verifiers or new code, you can choose SKIP during the Deduce Stage to move on to the next round.

\end{PromptBlock}

\paragraph{Nightmare - Question step - Not valid format prompt - OA}
\begin{PromptBlock}
You did not follow the required response format. Please try again with same choice.

**What You Must Do Now**:
- If you want to choose the next verifier to test, reply with verifier_num after <CHOICE> tag, such as <CHOICE>: 1.
- If you want to skip verifier testing for this round, reply with SKIP after <CHOICE> tag, such as <CHOICE>: SKIP.
- DO NOT include any explanation, only follow the response format.

**Response format**:
<CHOICE>: [your_choice]
\end{PromptBlock}

\paragraph{Nightmare - Question step - Not valid verifier choice prompt - OA}
\begin{PromptBlock}
You selected Verifier <{verifier_num}>, which is not a valid verifier number.

Please choose a valid verifier or SKIP to next stage.

**What You Must Do Now**:
- If you want to choose the next verifier to test, reply with verifier_num after <CHOICE> tag, such as <CHOICE>: 1.
- If you want to skip verifier testing for this round, reply with SKIP after <CHOICE> tag, such as <CHOICE>: SKIP.
- DO NOT include any explanation, only follow the response format.

**Response format**:
<CHOICE>: [your_choice]
\end{PromptBlock}

\paragraph{Nightmare - Question step - First question prompt - CoT}
\begin{PromptBlock}
You are now entering the **Verifier Questioning Stage** of this round.

**Current Verifiers**:
{verifier_descriptions}

**Stage Purpose**:
In this stage, you can test your proposed 3-digit code using verifiers. Each verifier checks one hidden criterion. Use the test results to gather information and refine your deduction.

**Verifier Rules Summary**:
- Each verifier has ONE secretly selected active criterion.
- Each verifier shows results for a different, hidden verifier (the mapping is randomized but fixed for the entire game).
- <PASS> means your code satisfies the active criterion of the secretly mapped verifier. <FAIL> means your code does not satisfy that criterion.
- Active rules do NOT overlap between verifiers.

**Your Goal in This Stage**:
- Choosing verifiers is optional; testing 0 verifiers is allowed. If you want to choose the verifier, you must choose verifiers **one at a time**. After each result, you may decide whether to test another. You may choose to test 0 to 3 verifiers **in total** during this round.
- **Passing all tested verifiers does NOT mean the code is correct.** To win, your code must satisfy the hidden criterion of **all verifiers**, whether tested or not.

**What You Must Do Now**:
- If you want to choose a verifier to test your proposed code, reply with verifier_num after <CHOICE> tag, such as <CHOICE>: 1.
- If you want to skip verifier testing for this round, reply with SKIP after <CHOICE> tag, such as <CHOICE>: SKIP.
- Explain your reasoning step by step based on verifier result after <REASONING> tag, then provide your choice.

**Response format**:
<REASONING>: [Explain your reasoning step by step for choosing the verifier or skipping verifiers]
<CHOICE>: [your_choice]
\end{PromptBlock}

\paragraph{Nightmare - Question step - Following questions prompt - CoT}
\begin{PromptBlock}
You chose Verifier <{verifier_num}> and the result is <{verifier_result}>.

**Hint**:
- `<PASS>` means your code satisfies the active criterion of the actual Verifier it is mapped to. For example, if Verifier 1 is secretly mapped to Verifier 2, then a <PASS> from Verifier 1 means your code met Verifier 2's hidden active rule.
- `<FAIL>` means your code does not satisfy the active criterion of the actual Verifier it is mapped to. 

**What You Must Do Now**:
- If you want to choose the next verifier to test, reply with verifier_num after <CHOICE> tag, such as <CHOICE>: 1.
- If you want to skip verifier testing for this round, reply with SKIP after <CHOICE> tag, such as <CHOICE>: SKIP.
- Explain your reasoning step by step based on verifier result after <REASONING> tag, then provide your choice.

**Response format**:
<REASONING>: [Explain your reasoning step by step for choosing the verifier or skipping verifiers]
<CHOICE>: [your_choice]
\end{PromptBlock}

\paragraph{Nightmare - Question step - Last question prompt - CoT}
\begin{PromptBlock}
You chose Verifier <{verifier_num}> and the result is <{verifier_result}>.

You have now tested the maximum number of three verifiers for this round. The next stage is the Deduce Stage. If you want to test more verifiers or new code, you can choose SKIP during the Deduce Stage to move on to the next round.
\end{PromptBlock}

\paragraph{Nightmare - Question step - Not valid format prompt - CoT}
\begin{PromptBlock}
You did not follow the required response format. Please try again with same choice.

**What You Must Do Now**:
- If you want to choose the next verifier to test, reply with verifier_num after <CHOICE> tag, such as <CHOICE>: 1.
- If you want to skip verifier testing for this round, reply with SKIP after <CHOICE> tag, such as <CHOICE>: SKIP.
- Explain your reasoning step by step based on verifier result after <REASONING> tag, then provide your choice.

**Response format**:
<REASONING>: [Explain your reasoning step by step for choosing the verifier or skipping verifiers]
<CHOICE>: [your_choice]
\end{PromptBlock}

\paragraph{Nightmare - Question step - Not valid verifier choice prompt - CoT}
\begin{PromptBlock}
You selected Verifier <{verifier_num}>, which is not a valid verifier number.

Please choose a valid verifier or SKIP to next stage.

**What You Must Do Now**:
- If you want to choose the next verifier to test, reply with verifier_num after <CHOICE> tag, such as <CHOICE>: 1.
- If you want to skip verifier testing for this round, reply with SKIP after <CHOICE> tag, such as <CHOICE>: SKIP.
- Explain your reasoning step by step based on verifier result after <REASONING> tag, then provide your choice.

**Response format**:
<REASONING>: [Explain your reasoning step by step for choosing the verifier or skipping verifiers]
<CHOICE>: [your_choice]
\end{PromptBlock}

\subsubsection{Nightmare deduce step prompt}

\paragraph{Nightmare - Deduce step - Deduce result prompt}
\begin{PromptBlock}
The final guess is {submitted_code}. The answer is {answer}, the guess is {is_correct}.
\end{PromptBlock}

\paragraph{Nightmare - Deduce step - Step prompt - OA}
\begin{PromptBlock}
You are now entering the **Deduce Stage** of this round.

**Stage Purpose**:
In this stage, you can analyze all the information gathered then decide whether to continue to the next round or submit a final guess.

**Hint**:
- Passing all tested verifiers does not mean the code is correct if not all verifiers were tested. To be correct, the code must satisfy the hidden criteria of all verifiers, not just the ones you tested.
- You may choose not to test some verifiers if you can clearly reason that your code meets their requirements. But you must ensure every verifier is either tested and passed, or clearly justified through reasoning. Testing and passing only part of the verifiers is not enough if others are ignored.
- This stage **is not for testing**, you don't have to submit an answer; you can proceed to the next round to continue gathering information.
- Accuracy takes priority over speed. If you submit, the game will end, and an incorrect guess will result in immediate failure.

**Your Goal in This Stage**:
- Decide whether to submit the final guess or continue to the next round. Submit the final guess will end the game, continue to the next round will help you gather more information.
- Submission is not mandatory, you must make this decision based on your own reasoning.

**What You Must Do Now**:
- If you want to continue to the next round, reply with SKIP after <CHOICE> tag, such as <CHOICE>: SKIP
- If you want to submit a final guess to end the game, reply with BLUE=X, YELLOW=Y, PURPLE=Z after <CHOICE> tag, such as <CHOICE>: BLUE=1, YELLOW=1, PURPLE=1.
- DO NOT include any explanation, only follow the response format.

**Response format**:
<CHOICE>: [your_choice]
\end{PromptBlock}

\paragraph{Nightmare - Deduce step - Not valid format prompt - OA}
\begin{PromptBlock}
You did NOT follow the response format. Please try again.

**What You Must Do Now**:
- If you want to continue to the next round, reply with SKIP after <CHOICE> tag, such as <CHOICE>: SKIP
- If you want to submit a final guess to end the game, reply with BLUE=X, YELLOW=Y, PURPLE=Z after <CHOICE> tag, such as <CHOICE>: BLUE=1, YELLOW=1, PURPLE=1.
- DO NOT include any explanation, only follow the response format.

**Response format**:
<CHOICE>: [your_choice]
\end{PromptBlock}

\paragraph{Nightmare - Deduce step - Step prompt - CoT}
\begin{PromptBlock}
You are now entering the **Deduce Stage** of this round.

**Stage Purpose**:
In this stage, you can analyze all the information gathered then decide whether to submit a final guess or continue to the next round.

**Hint**:
- Passing all tested verifiers does not mean the code is correct if not all verifiers were tested. To be correct, the code must satisfy the hidden criteria of all verifiers, not just the ones you tested.
- You may choose not to test some verifiers if you can clearly reason that your code meets their requirements. But you must ensure every verifier is either tested and passed, or clearly justified through reasoning. Testing and passing only part of the verifiers is not enough if others are ignored.
- This stage **is not for testing**, you don't have to submit an answer; you can proceed to the next round to continue gathering information.
- Accuracy takes priority over speed. If you submit, the game will end, and an incorrect guess will result in immediate failure.

**Your Goal in This Stage**:
- Analysis all information gathered.
- Decide whether to submit the final guess or continue to the next round.
- Submission is not mandatory, you must make this decision based on your own reasoning.

**What You Must Do Now**:
- If you want to continue to the next round, reply with SKIP after <CHOICE> tag, such as <CHOICE>: SKIP
- If you want to submit a final guess to end the game, reply with BLUE=X, YELLOW=Y, PURPLE=Z after <CHOICE> tag, such as <CHOICE>: BLUE=1, YELLOW=1, PURPLE=1.
- Explain your reasoning step by step with <REASONING> tag, then provide your choice. If you want to submit a final guess, you must provide the reasons for not proceeding to the next round.

**Response format**:
<REASONING>: [Analysis and explain your reasoning step by step for continue to next round or submit final guess]
<CHOICE>: [your_choice]
\end{PromptBlock}

\paragraph{Nightmare - Deduce step - Not valid format prompt - CoT}
\begin{PromptBlock}
You did NOT follow the response format. Please try again.

**What You Must Do Now**:
- If you want to continue to the next round, reply with SKIP after <CHOICE> tag, such as <CHOICE>: SKIP
- If you want to submit a final guess to end the game, reply with BLUE=X, YELLOW=Y, PURPLE=Z after <CHOICE> tag, such as <CHOICE>: BLUE=1, YELLOW=1, PURPLE=1.
- Explain your reasoning step by step with <REASONING> tag, then provide your choice.

**Response format**:
<REASONING>: [Analysis and explain your reasoning step by step for submitting the final guess or continue to next round]
<CHOICE>: [your_choice]
\end{PromptBlock}

\section{Human Player Interface}
\label{apx:human_player_interface}

\begin{figure}[ht]
  \centering
  \includegraphics[width=\textwidth]{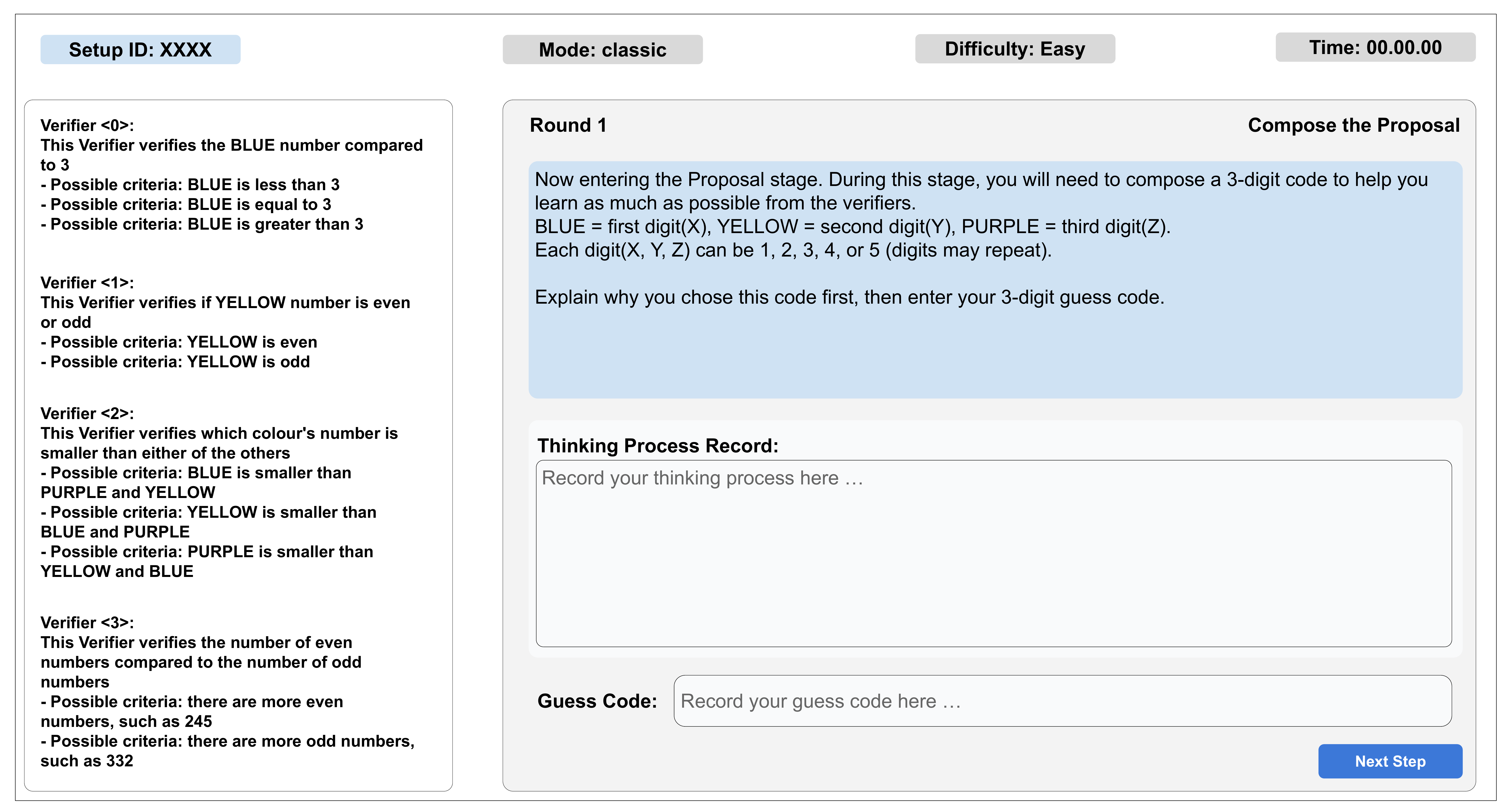}
  \caption{The interface for a human player in the game, providing exactly the same information as an LLM player. The top section displays the basic information of the current game, the left side shows the verifier information, and the right side includes stage introduction and the interaction panel. The human player first needs to think according to the requirements of the round and record their reasoning process, then provide their decision, and finally click "next step" to move on to the next phase. This process is fully consistent with the LLM player’s flow.}
  \label{fig:human_interface}
\end{figure}

\end{document}